\pdfoutput=1

\documentclass[11pt]{article}

\usepackage[]{acl}

\usepackage{times}
\usepackage{latexsym}

\usepackage[T1]{fontenc}

\usepackage[utf8]{inputenc}

\usepackage{microtype}

\usepackage{hyperref}       
\usepackage{url}            
\usepackage{booktabs}       
\usepackage{amsfonts}       
\usepackage{nicefrac}       
\usepackage{graphicx}       
\usepackage{natbib}       
\usepackage{amsmath, amssymb, amsfonts, bm, mathtools}       

\usepackage{makecell}
\usepackage{multirow}
\usepackage{subfigure}
\usepackage{enumitem}
\usepackage{tabularx}
\usepackage{xspace}
\usepackage{tcolorbox}
\usepackage{adjustbox}

\newcommand{\secref}[2][]{Section#1~\ref{sec:#2}}

\newcommand{\tabref}[2][]{Table#1~\ref{tab:#2}}
\newcommand{\figref}[2][]{Figure#1~\ref{fig:#2}}

\newcommand{\appref}[2][]{Appendix#1~\ref{#2}}

\usepackage{longtable}
\definecolor{DeepSkyBlue}{RGB}{0,191,255}
\definecolor{darkpastelred}{rgb}{0.76, 0.23, 0.13}
\definecolor{darkpastelgreen}{rgb}{0.01, 0.75, 0.24}
\definecolor{Mulberry}{rgb}{0.77,0.29,0.55}
\definecolor{CadmiumOrange}{rgb}{0.93,0.53, 0.18}
\definecolor{ForestGreen}{rgb}{0.13, 0.55, 0.13}
\definecolor{WildStrawberry}{rgb}{0.5, 0.7, 0.2}

\title{Decomposed Opinion Summarization with Verified Aspect-Aware Modules}

\author{Miao Li$^{1,3}$ \quad Jey Han Lau$^1$ \quad Eduard Hovy$^{1, 2}$ \quad Mirella Lapata$^3$ \\
       $^1$School of Computing and Information Systems, The University of Melbourne \\
       $^2$Language Technologies Institute, Carnegie Mellon University\\
       $^3$School of Informatics, The University of Edinburgh\\
       \texttt{miao4@student.unimelb.edu.au},\\ \texttt{\{laujh, eduard.hovy\}@unimelb.edu.au},\\ \texttt{mlap@inf.ed.ac.uk}
       }

\begin{document}
\maketitle
\begin{abstract}
Opinion summarization plays a key role in deriving meaningful insights from large-scale online reviews.
To make the process more explainable and grounded, we propose a domain-agnostic modular approach guided by review aspects (e.g., cleanliness for hotel reviews) which separates the tasks of aspect identification, opinion consolidation, and meta-review synthesis to enable greater transparency and ease of inspection.
We conduct extensive experiments across datasets representing scientific research, business, and product domains. 
Results show that our approach generates more grounded summaries compared to strong baseline models, as verified through automated and human evaluations. 
Additionally, our modular approach, which incorporates reasoning based on review aspects, produces more informative intermediate outputs than other knowledge-agnostic decomposition approaches.
Lastly, we provide empirical results to show that these intermediate outputs can support humans in summarizing opinions from large volumes of reviews.\footnote{Code and data are released at: \url{https://github.com/oaimli/ModularMetaReview}}
\end{abstract}

\section{Introduction}

Reviews are omnipresent in the digital world, providing invaluable insights into products~\citep{amasum_2021}, businesses~\citep{space_2021}, even scientific articles~\citep{peersum_2023}. Automatic opinion summarization aims to \emph{aggregate} a large and diverse set of reviews about a particular \emph{entity} (e.g., hotel) into a single easy-to-read \emph{meta-review} (or summary). 
A good meta-review should accurately reflect the balance of opinions in the source reviews and speak to the entity's specific \emph{aspects} (e.g., \emph{Cleanliness}, \emph{Service}, \emph{Location}). 
A useful meta-review should also present some \emph{evidence} justifying its content.

Opinion summarization has distinct characteristics that set it apart from other summarization tasks. 
Firstly, it cannot rely on reference summaries for training, as human-written meta-reviews are not generally available (e.g.,~across entities and domains) and can be difficult to crowdsource (e.g.,~for entities represented by thousands of reviews). 
Secondly, a meta-review needs to cover the most important aspects related to the entity of interest.
And finally, given the subjective nature of the summarization task, systems should offer some evidence to justify their output.

Prior approaches to generating meta-reviews broadly fall into three categories. \emph{Extractive} methods create summaries by selecting a few representative sentences from source reviews~\citep{space_2021, basu-roy-chowdhury-etal-2022-unsupervised, aspect_aware_extractive_summarization_2023}. While these approaches are scalable and inherently attributable, the summaries tend to be overly detailed and lack coherence. 
\emph{Abstractive} methods rely on neural language models to generate fluent and coherent meta-reviews with novel language~\citep{document_structure_aspect_summarization_2019,meansum_2019,coavoux-etal-2019-unsupervised,brazinskas-etal-2020-unsupervised,amplayo-etal-2021-aspect,amplayo2021unsupervised,iso-etal-2021-convex-aggregation,amasum_2021,cattan-etal-2023-key}. In the era of large language models (LLMs), long-context language models could be directly used on opinion summarization with prompting~\citep{llama_2023, gpt4_2023}.
However, these abstractive approaches are neither
transparent nor controllable due to the black-box nature of
end-to-end modeling.

\emph{Hybrid} methods
\cite{attributable_opinion_summarization_2023,prompted_opinion_summarization_2023,hiro_2024,meta_review_logic_2024}, the third category of summarisation approaches, could generate fluent and explainable summaries. They first extract information clusters in the format of sentence fragments~\citep{attributable_opinion_summarization_2023,prompted_opinion_summarization_2023,hiro_2024} or sentiment spans~\citep{meta_review_logic_2024}, and then generate summaries (e.g., using a language model) based on the clusters. However, they are limited in that they are based on assumptions valid for specific domains (e.g.,~based on the popularity of opinions) and are not entirely transparent (e.g.,~either the clusters or aggregation step cannot be easily verified).  


\begin{figure*}[t]
\centering
\includegraphics[width=0.98\textwidth, trim=63 350 75 270, clip]{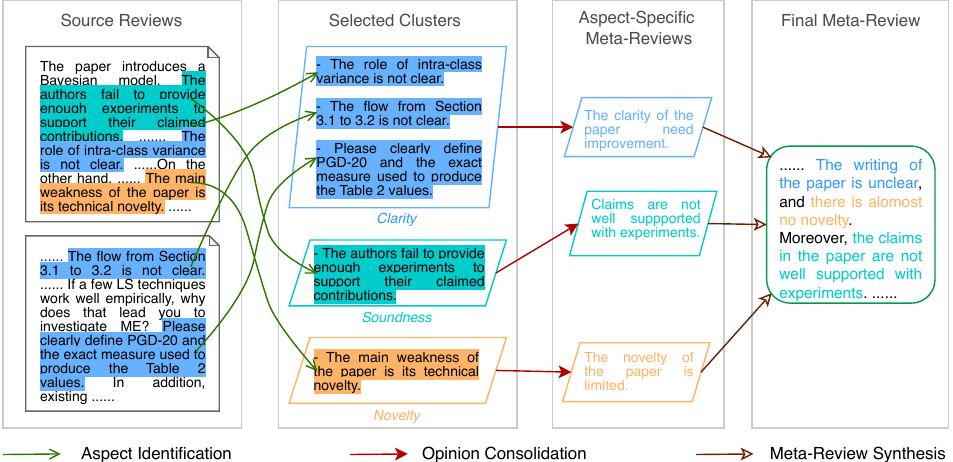}
\caption{High-level overview of our decomposition for opinion
  summarization using an example from the scientific domain with three
  aspects ({\color{blue}{\textit{Clarity}}},
  {\color{cyan}{\textit{Soundness}}}, and
  {\color{orange}{\textit{Novelty}}}). The modules \textit{Aspect
    Identification}, \textit{Opinion Consolidation}, and
  \textit{Meta-Review Synthesis} are instantiated with prompt-based
  LLMs and operate in sequence. The output of \textit{Aspect Identification}
  serves as input to \textit{Opinion consolidation} and \textit{Meta-Review synthesis}
  aggregates opinions found in aspect-specific meta-reviews. All prompts
  and inputs/outputs are in natural language.}
\label{fig:modules}
\vspace{-1em}
\end{figure*}

In this paper, we propose to decompose opinion summarization into simpler sub-tasks that are domain-agnostic and can be executed by prompting-based LLMs dedicated to these sub-tasks. Our approach is also inspired by recent applications of
chain-of-thought prompting \cite{cot_2022} and its variants
\cite{decomposed_prompting_2023,zhou2023leasttomost}, which address
reasoning problems by decomposing complex tasks into a sequence of
simpler sub-problems, which are solved sequentially. Our decomposition
consists of three high-level modules, namely \emph{Aspect
Identification}, \emph{Opinion Consolidation}, and \emph{Meta-Review
Synthesis}.  Intuitively, we first identify text fragments in the input
reviews discussing aspects about the entity and domain in
question; next we create meta-reviews for \emph{each} aspect, and
finally we generate a global meta-review for \emph{all} aspects (see
Figure~\ref{fig:modules}). Our approach eschews problems relating to
the scale of the input to some extent, since reviews can be processed in parallel to
identify their aspects. It also avoids problems with clusters being
diffuse or irrelevant since we leverage domain specific aspect
definitions (as part of the prompt) to obtain interpretable
clusters. Finally, our decomposition is controllable, and
evidence-based, as the output of each module can be traced back to its
input.
Our contributions can be summarized as follows:
\begin{itemize}
    \item We propose a domain-agnostic decomposition of opinion summarization into
      three verifiable modules that are instantiated with LLMs using zero-shot prompting.

      
    \item Extensive experiments on three datasets from different
      domains demonstrate that our aspect-informed approach produces
      more grounded meta-reviews than strong baselines in terms of
      automatic and human evaluation.
      
    \item Aspect-aware
      decomposition yields more useful reasoning chains compared to end-to-end prompting with automatic decomposition
      \cite{decomposed_prompting_2023}, and our experiments show that our generated intermediate reasoning steps are empirically helpful in assisting humans with summarizing reviews. 
\end{itemize}

\section{Related Work}


Our work focuses on abstractive opinion summarization that aims to
generate fluent and coherent summaries with novel
language~\cite{amasum_2021, peersum_2023}.
This task has been explored in different domains, such as summarizing
reviews of products, businesses, and scientific
articles~\citep{meansum_2019, amasum_2021, peersum_2023, hiro_2024}.
Previous abstractive methods lack transparency in their decision-making process
due to their end-to-end nature~\citep{amasum_2021,cattan-etal-2023-key, llama_2023, gpt4_2023}. Hybrid approaches implement pipelines
with transparent intermediate outputs, however, most of them are aspect
agnostic, focusing on how to organize or annotate the input for
downstream processing. For example, \citet{hiro_2024} propose a method that represents
sentences from reviews as paths through a learned discrete hierarchy,
and then use LLMs to generate sentences based on frequent paths
retrieved from this hierarchy. Their retrieval module relies
on majority voting, which is less effective in domains where minority but well-argued
opinions are valuable, such as in scientific reviews~\citep{peersum_2023}.

Inspired by a well-known decomposition of
multi-document summarization into three modules, namely content selection, content consolidation
(or fusion), and output generation~\citep{sentence_fusion_2005,
  multiple_online_sources_1998,lebanoff-etal-2020-cascade,slobodkin-etal-2024-multi,soap_2021,meta_review_logic_2024}, 
we apply a similar aspect-guided decomposition to the task of opinion summarization. A few approaches take aspects into account. For example, \citet{space_2021} cluster opinions through a discrete latent variable model and extract sentences based on popular aspects or a particular aspect, while \citet{aspect_aware_extractive_summarization_2023} learn aspects by clustering of aspect-related words. \citet{amplayo-etal-2021-aspect} fine-tune pre-trained models on synthetic data enhanced with aspect annotations which can be used to control output summaries at inference time. Different from earlier work \citep{aspect_aware_extractive_summarization_2023, prompted_opinion_summarization_2023, attributable_opinion_summarization_2023} which identifies aspects based on sentences, our approach identifies opinions in text fragments of variable lengths. We delegate the task of aspect identification to prompt
engineering, demonstrating that LLMs can reliably extract aspects in the format of flexible text fragments given an input review and aspect definitions without additional training.
\citet{prompted_opinion_summarization_2023} and \citet{meta_review_logic_2024} use similar modules to ours. However, intermediate results of their recursive prompting and aspect identification
are not inspected or verified and there is limited transparency, and \citet{meta_review_logic_2024} focus exclusively aspects in scientific reviews 
(e.g.,~opinions about \emph{Novelty} or \emph{Soundness}).

%



Our work also relates to recent efforts aiming to improve the in-context
learning performance of LLMs through intermediate reasoning
chains~\citep{cot_2022, tot_2023, decomposed_prompting_2023}. Previous
approaches focus primarily on mathematical or symbolic reasoning,
while intermediate reasoning for complex writing tasks such as opinion
summarization remains under-explored~\citep{meta_review_logic_2024}.
Decomposed prompting~\citep{decomposed_prompting_2023} uses LLMs to predict
both the task decomposition into modules and the modules
themselves. However, it is unclear whether it is  suited to complex tasks like opinion summarization.


%

\section{Task Decomposition}
\label{sec:task_decomposition}

Let $\mathcal{C}$ denote a corpus of reviews on entities $\{e_1, e_2, \dots\}$
from a domain $d$, for example, hotels or scientific articles.  Reviews
may discuss a number of relevant aspects $A_d = \{a_1,a_2,\dots\}$,
like \emph{Clarity} or \emph{Soundness}, For each entity $e_i$, our
task is to generate a meta-review $\widehat{y}_i$ by synthesizing
opinions from a set of source reviews \mbox{$R_i=\{r_1, r_2, \dots\}$}
covering \emph{all} attested aspects $A_d$.  We decompose the task
into three modules, namely \textit{Aspect Identification},
\textit{Opinion Consolidation}, and \textit{Meta-Review Synthesis}. We
present the inner workings of each module in~\figref{modules} with an
example from the scientific domain. Due to the limited availability of
training data, we implement our modules using an unsupervised
approach, leveraging zero-shot prompting of LLMs and their instruction following and generation
capabilities.\footnote{It is worth noting that our prompts could be
further improved, however, we leave prompt optimization to future
work.}


\paragraph{Aspect Identification}
As not all content in the source reviews is relevant for generating
meta-reviews, opinion summarization models must be able to isolate
critical information in the input.  The first module, \textit{Aspect
  Identification}, selects text fragments of variable lengths from
source reviews discussing any review aspect. Specifically, for
reviewed entity $e_i$, our module identifies text fragments for
aspect~$a_j$ from source reviews $R_i$. The module essentially
partitions text fragments into aspect-specific clusters
$C_{i, j}=\{f_1, f_2, \dots\}$, where fragments~$f_m$ can originate
from any source review in $R_i$. For example, in \figref{modules}, the module
identifies fragments in scientific reviews for the aspects
\textit{Clarity}, \textit{Soundness}, and \textit{Novelty}. We
implement this module with zero-shot LLM prompting. Our prompt
template is shown in
Appendix~\ref{sec:appendix_prompts_modular_approach}
(Figure~\ref{tab:prompt_extraction}) and can be modified for
different aspects and domains. Our aspect identification is not based on sentence clustering as well-justified opinions may be composed of multiple sentences, and each text fragment could be identified for multiple aspects.

\paragraph{Opinion Consolidation} As shown in
Figure~\ref{fig:modules}, the output of the first module
consists of clusters of text fragments, each discussing a specific
aspect. Depending on the domain, these clusters can have a lot of
redundancy, often repeating the same opinion. Our second module,
\textit{Opinion Consolidation}, aggregates opinions into
aspect-specific meta-reviews. We essentially adopt a divide-and-conquer
strategy, since generating meta-reviews from aspect-specific clusters
is significantly easier than producing an entire summary from reviews
containing mixed aspects.  Specifically, taking as input cluster
$C_{i, j}$, the module generates meta-review~$o_{i, j}$ for
aspect~$a_{j}$.  As we do not have training data for these
intermediate summaries, we also implement this module with zero-shot
prompting.\footnote{Some aspects may not have
 corresponding text fragments in the source reviews, as they do 
not always cover every aspect.}  Our template (shown in the Appendix,
Figure~\ref{tab:prompt_reasoning}) instructs LLMs to integrate
opinions (i.e.,~text fragments) from a specific cluster. For example,
in \figref{modules}, the three sentences in the \textit{Clarity}
cluster are aggregated into ``\textsl{The clarity of the paper needs
  improvement}''.

\paragraph{Meta-Review Synthesis}

After obtaining all aspect-specific summaries $O_{i} = \{o_{i,1}, o_{i,2},
\dots\}$, our last module generates the final meta-review
$\widehat{y}_i$ for entity~$e_i$; it combines the opinions mentioned
in the individual summaries into a fluent and coherent overall
summary. An example is given in ~\figref{modules} where the
meta-review focuses on the aspects of \textit{Clarity},
\textit{Soundness}, and \textit{Novelty}. Again, this module leverages
the generation capabilities of LLMs, and is instantiated via zero-shot
prompting. Our template (given in the Appendix,
Figure~\ref{tab:prompt_generation}) asks the LLM to write a concise
meta-review which summarizes the provided opinions and covers all
mentioned aspects.  

\section{Experimental Setup}

We showcase the versatility of our approach on different domains. We first describe the datasets used in our experiments, discuss
implementation details and comparison baselines, and explain how we
evaluate performance with automatic metrics.

\paragraph{Datasets}
\label{sec:datasets}

We conducted experiments on three domains, product reviews for sports
shoes, business reviews for hotels, and scientific reviews for
research articles. For business reviews, we use SPACE, an opinion
summarization dataset constructed by~\citet{space_2021}. For product
reviews, we use the sports shoes subset from
AmaSum~\citep{amasum_2021}. For scientific reviews, we use
PeerSum~\citep{peersum_2023} and also the human annotations of review
aspects from~\citet{meta_review_logic_2024}.\footnote{To make our experiments cost-effective, we randomly sampled 100 test instances from the PeerSum dataset.} Statistics for these
datasets are shown in Table~\ref{tab:data_statistics}. 

SPACE~\citep{space_2021} consists of hotel reviews from TripAdvisor,
 with 100 reviews per entity, as well as reference meta-reviews of
 customer experiences created by annotators. The dataset provides six
 aspects for hotels, which we adopt in our experiments, namely
 \textit{Building}, \textit{Cleanliness}, \textit{Food},
 \textit{Location}, \textit{Rooms}, and \textit{Service}. AmaSum
 contains meta-reviews for a variety of Amazon products, with reference summaries collated
 from professional review platforms. To cover more domains in our experiments with limited computing resources, we randomly choose the sports shoes
 subset curated from the RunRepeat platform\footnote{\scriptsize https://runrepeat.com/} which covers the aspects: 
 \textit{Breathability}, \textit{Durability}, \textit{Weight},
 \textit{Cushioning}, \textit{Stability}, \textit{Flexibility},
 \textit{Traction}, \textit{Size and Fit}, \textit{Comfort}, and
 \textit{Misc}. 
 
 PeerSum \cite{meta_review_logic_2024} contains reviews
 for scientific articles and corresponding meta-reviews from OpenReview
 focusing on the aspects of \textit{Novelty}, \textit{Soundness},
 \textit{Clarity}, \textit{Advancement}, and
 \textit{Compliance}. Detailed definitions for all aspects in the datasets (SPACE,
 AmaSum, and PeerSum) are given in the Appendix
\ref{sec:prompts_scientific}--\ref{sec:prompts_product}. 

\begin{table}[t]
    \setlength\tabcolsep{3pt}
\centering
\begin{adjustbox}{max width=\linewidth}
    \begin{tabular}{@{}l@{~}r@{~}r@{~}r@{~}r@{~}r@{}}
    \toprule
    \textbf{Dataset} & \textbf{\#Train}/ \textbf{Dev}/\textbf{Test} & \textbf{\#Reviews} & \textbf{SourceL} & \textbf{MetaL} & \textbf{\#Aspects}\\
    \midrule
    PeerSum & 22,420/50/100 & \hspace{.5ex}14.9 & \hspace{.5ex}5,146 & 156.1 & 5\\
    AmaSum & 25,203/50/50 & 381.8 & 14,495 & 94.8 & 10\\
    SPACE & 0/25/25 & 100\hspace{.25cm} & 14,439 & 75.7 & 6\\
    \bottomrule
    \end{tabular}
    \end{adjustbox}
    \caption{Statistics of our experimental datasets. \#Train/Dev/Test
      refer to the number of training, development, and test
      instances, respectively; \#Reviews is the average number of
      reviews per entity; SourceL refers to the total length of the source
      reviews (when concatenated) and MetaL to the average meta-review
      length; \#Aspects is the number of aspects covered in each
      dataset. For AmaSum, the statistics are for the sports shoes
      subset.}
    \label{tab:data_statistics}
    \vspace{-1em}
\end{table}

\begin{table*}[t]
     \small
    \setlength\tabcolsep{2pt}
    \centering
    \begin{adjustbox}{max width=\linewidth}
    \begin{tabular}{lcccc}
    \toprule
    \multicolumn{1}{c}{\textbf{Models}} & \textbf{Coverage}$\uparrow$ & \textbf{G-Eval}$\uparrow$ & \textbf{AlignScore-R/M}$\uparrow$ & \textbf{Rouge}$\uparrow$ \\
    \midrule
    Sentiment CoT-GPT-4o~\citep{meta_review_logic_2024} & \underline{0.96} & 0.75 & \underline{0.72}/\underline{0.08} & \underline{\textbf{23.47}} \\
    FT-Llama 8B~\cite{llama_2023} & 0.87 & 0.60 & 0.33/0.06 & 20.60\\
    Aspect-aware decomposition-GPT-4o (ours) & 0.95 & \underline{0.76} & 0.68/0.06 & 20.78\\
    \midrule
    Automatic decomposition-Llama 8B~\citep{decomposed_prompting_2023} & 0.58 & 0.20 & 0.36/0.03 & 11.98\\
    Chunk-wise decomposition-Llama 8B~\citep{decomposed_prompting_2023} & 0.79 & 0.65 & 0.65/0.03 & \underline{21.19}\\
    Naive aspect-aware prompting-Llama 8B~\citep{gpt2_2019} & 0.72 & 0.62 & 0.70/0.06 & 16.93\\
    Aspect-aware decomposition-Llama 8B (ours) & \underline{0.90} & \underline{0.66} & \underline{0.71}/\underline{0.07} & 21.12\\
    \midrule
    Automatic decomposition-Llama 70B~\citep{decomposed_prompting_2023} & 0.59 & 0.31  & 0.51/0.03 & 12.0\\
    Chunk-wise decomposition-Llama 70B~\citep{decomposed_prompting_2023} & 0.84 & 0.72 & 0.65/0.06 & 21.80\\
    Naive aspect-aware prompting-Llama 70B~\citep{gpt2_2019} & 0.72 & 0.62 & 0.70/0.07 & 16.82\\
    Aspect-aware decomposition-Llama 70B  (ours) & \underline{\textbf{0.97}} & \underline{\textbf{0.76}} & \underline{\textbf{0.76}}/\underline{\textbf{0.09}} & \underline{22.58}\\
    \bottomrule
    \end{tabular}
    \end{adjustbox}
    \caption{Results on scientific \textbf{reviews of research
        articles}. The first section of the table presents results for
      GPT-4o and state-of-the-art models. The second section has
      results for Llama-8B, and the third one for
      Llama~70B. Underlined scores denote best in section per metric
      while bold scores denote best overall. AlignScore-R calculates
      AlignScore against source reviews, while AlignScore-M is computed
      against reference meta-reviews.}
    \label{tab:main_results_peersum}
\end{table*}

\paragraph{Model Comparisons}
We implement our modular approach with different backbone LLMs,
including closed- and open-source models. Since the modules need to
have reasonable language generation and instruction following
capabilities, we conduct experiments with
gpt-4o-2024-05-13\footnote{\scriptsize
https://platform.openai.com/docs/models/gpt-4o} from OpenAI,
and Llama-3.1-70B-Instruct\footnote{\scriptsize
https://huggingface.co/meta-llama/Llama-3.1-70B-Instruct} and Llama-3.1-8B-Instruct\footnote{\scriptsize
https://huggingface.co/meta-llama/Llama-3.1-8B-Instruct} from
Meta.\footnote{All models used in our experiments are
instruction-tuned.}  The prompts used in our experiments are provided
in Appendix~\ref{sec:prompts_scientific}--\ref{sec:prompts_product}.

We compare our approach with representative prompting and fine-tuning
baselines (see more details in~\appref{sec:implementation_baselines}). We implement two strong prompting approaches which do not
take aspect information into account: \emph{automatic decomposition}
breaks down complex reasoning tasks into simpler
ones~\citep{decomposed_prompting_2023} by automatically predicting the
decomposition and the modules, while \emph{chunk-wise decomposition}
\citep{decomposed_prompting_2023} recursively summarizes the input
reviews chunk-by-chunk with prompting.\footnote{The input is chunked based on
document boundaries. For PeerSum each review is a chunk, while for
AmaSum and SPACE chunks correspond to 20\% of the source documents.}
We also compare against the \emph{naive aspect-aware prompting} which
does not perform task decomposition but is aspect-aware~\citep{gpt2_2019}.
For
fine-tuning, we conduct experiments on decoder-only LLMs. Due to
computational limitations, we present fine-tuning results only with
Llama-3.1-8B\footnote{\scriptsize
https://huggingface.co/meta-llama/Llama-3.1-8B} on all three
datasets. Moreover, we also include generations from strong baseline
approaches on our datasets.


\paragraph{Automatic Evaluation Metrics}
We evaluate the quality of generated meta-reviews in terms of
\emph{aspect coverage} and \emph{faithfulness} (against source
reviews). Aspect coverage measures how well the generated meta-review
for entity~$e_i$ captures the aspects discussed in the source
reviews. Specifically, we compute the $F_1$ between the set of aspects
present in the generated meta-review and those in the source
reviews. We recognize aspects automatically by running our Aspect
Identification module (see Section~\ref{sec:task_decomposition}) on
the system input and output.  Opinion faithfulness measures how well
opinions in generated meta-reviews are supported by the
source reviews. Specifically, we use G-Eval~\citep{geval_2021}, a
prompting-based evaluation\footnote{Our prompts are provided
in~\appref{sec:implementation_metrics}.} metric, and
AlignScore~\citep{alignscore_2023}\footnote{\scriptsize
https://github.com/yuh-zha/AlignScore/tree/main}, a fine-tuned
evaluation metric based on information alignment
between two arbitrary text pieces. We use the large version of the
pre-trained backbone for AlignScore, and we set \textit{nli\_sp} as
our evaluation mode.
We also report Rouge~F1~\citep{rouge_2003}, as a measure of overall
summary quality.\footnote{We use the average F1 of ROUGE-1, ROUGE-2, and ROUGE-L.} To obtain fair conclusions, we make the models output three generations for each instance and present the average performance in the tables.

\begin{table*}[t]
     \small
    \setlength\tabcolsep{2pt}
    \centering
    \begin{adjustbox}{max width=\linewidth}
    \begin{tabular}{lcccc}
    \toprule
    \multicolumn{1}{c}{\textbf{Models}} & \textbf{Coverage}$\uparrow$ & \textbf{G-Eval}$\uparrow$ & \textbf{AlignScore-R/M}$\uparrow$ & \textbf{Rouge}$\uparrow$\\
    \midrule
    HIRO-abs~\citep{hiro_2024} & 0.54 & 0.35 & 0.78/0.13 & 14.90\\
    FT-Llama 8B~\citep{llama_2023} & 0.45 & 0.12 & 0.43/0.16 & 9.90\\ 
    Aspect-aware decomposition-GPT-4o (ours) & \underline{\textbf{0.86}} & \underline{0.87} & \underline{\textbf{0.79}}/\underline{\textbf{0.17}} & \underline{16.10}\\
    \midrule
    Automatic decomposition-Llama 8B~\citep{decomposed_prompting_2023} & 0.39 & 0.11 & 0.47/\underline{0.13} & 9.23\\
    Chunk-wise decomposition-Llama 8B~\citep{decomposed_prompting_2023} & 0.58 & \underline{0.80} & 0.66/0.08 & \underline{\textbf{16.59}}\\
    Naive aspect-aware prompting-Llama 8B~\citep{gpt2_2019} & 0.54 & 0.29 & 0.50/0.07 & 8.80\\
    Aspect-aware decomposition-Llama 8B (ours) & \underline{0.77} & 0.78 & \underline{0.69}/0.09 & 16.44\\
    \midrule
    Automatic decomposition-Llama 70B~\citep{decomposed_prompting_2023} & 0.31 & 0.28 & 0.68/0.14 & 7.74\\
    Chunk-wise decomposition-Llama 70B~\citep{decomposed_prompting_2023} & 0.57 & \underline{\textbf{0.88}} & 0.54/0.07 & 15.28\\
    Naive aspect-aware prompting-Llama 70B~\citep{gpt2_2019} & 0.49 & 0.48 & 0.60/0.09 & 7.35\\
    Aspect-aware decomposition-Llama 70B (ours) & \underline{0.83} & 0.86 & \underline{0.74}/\underline{0.16} & \underline{16.40}\\
    \bottomrule
    \end{tabular}
    \end{adjustbox}
    \caption{Results on product \textbf{reviews of sports shoes}.
 The first section of the table presents results for
      GPT-4o and state-of-the-art models. The second section has
      results for Llama-8B, and the third one for
      Llama~70B. Underlined scores denote best in section per metric
      while bold scores denote best overall. AlignScore-R calculates
      AlignScore against source reviews, while AlignScore-M is computed
      against reference meta-reviews.}
    \label{tab:main_results_amasum}
\end{table*}

\begin{table*}[t]
     \small
    \setlength\tabcolsep{2pt}
    \centering
    \begin{adjustbox}{max width=\linewidth}
    \begin{tabular}{lcccc}
    \toprule
    \multicolumn{1}{c}{\textbf{Models}} & \textbf{Coverage}$\uparrow$ & \textbf{G-Eval}$\uparrow$ & \textbf{AlignScore-R/M}$\uparrow$ & \textbf{Rouge}$\uparrow$\\
    \midrule
    HIRO-abs~\citep{hiro_2024} & 0.87 & 0.62 & \underline{\textbf{0.83}}/\underline{\textbf{0.24}} & \underline{\textbf{26.50}}\\
    TCG~\citep{prompted_opinion_summarization_2023} & 0.98 & 0.66 & 0.66/0.11 & 22.98\\
    Aspect-aware decomposition-GPT-4o (ours) & \underline{\textbf{1.00}} & \underline{\textbf{0.90}} & 0.81/0.10 & 21.38\\
    \midrule
    Automatic decomposition-Llama 8B~\citep{decomposed_prompting_2023} & 0.65 & 0.07 & 0.55/0.15 & 13.80\\
    Chunk-wise decomposition-Llama 8B~\citep{decomposed_prompting_2023} & 0.94 & 0.80 & 0.65/0.14 & \underline{22.9}\\
    Naive aspect-aware prompting-Llama 8B~\citep{gpt2_2019} & 0.55 & 0.06 & 0.34/\underline{0.18} & 10.30\\
    Aspect-aware decomposition-Llama 8B (ours) & \underline{0.97} & \underline{0.81} & \underline{0.70}/0.10 & 22.05\\
    \midrule
    Automatic decomposition-Llama 70B~\citep{decomposed_prompting_2023} & 0.63 & 0.38 & 0.70/\underline{0.22} & 10.0\\
    Chunk-wise decomposition-Llama 70B~\citep{decomposed_prompting_2023} & 0.93 & 0.84 & 0.65/0.01 & 22.02\\
    Naive aspect-aware prompting-Llama 70B~\citep{gpt2_2019} & 0.37 & 0.34 & 0.44/0.22 & 5.00\\
    Aspect-aware decomposition-Llama 70B (ours) & \underline{0.99} & \underline{0.88} & \underline{0.79}/0.11 & \underline{23.46}\\
    \bottomrule
    \end{tabular}
    \end{adjustbox}
    \caption{Results on business \textbf{reviews of
        hotels}.  The first section of the table presents results for
      GPT-4o and state-of-the-art models. The second section has
      results for Llama-8B, and the third one for
      Llama~70B. Underlined scores denote best in section per metric
      while bold scores denote best overall. AlignScore-R calculates
      AlignScore against source reviews, while AlignScore-M is computed
      against reference meta-reviews.}
    \label{tab:main_results_space}
    \vspace{-1.5em}
\end{table*}

\section{Results and Analysis}
\label{sec:results_and_analysis}

We perform experiments on datasets covering multiple domains,
comparing meta-reviews generated by our approach with those from
strong baselines and state-of-the-art approaches.  We further evaluate
the intermediate outputs obtained from our modules against human
annotations and conduct ablations to examine the extent to which
individual modules contribute to the summarization task. Finally, in
addition to automatic evaluation we conduct human evaluation based on
pair-wise system comparisons and intermediate outputs. 


\paragraph{Aspect-aware decomposition leads to better aspect coverage
  and opinion faithfulness.}
Our results using automatic evaluation metrics are summarized
in~\tabref{main_results_peersum} (scientific articles),
\tabref{main_results_amasum} (shoes), and~\tabref{main_results_space}
(hotels).\footnote{We run inference three times, with different random
seeds and report average performance.}  Across domains we find that
our modular approach with GPT-4o or Llama-3.1-70B delivers the highest
coverage of review aspects. Our approach with GPT-4o is also better
than comparison systems in terms of opinion faithfulness (see
AlignScore).
Our aspect-aware decomposition is consistently superior
to more naive decompositions and prompting methods in terms of
aspect coverage across domains and model backbones. We also observe that
using Llama-70B as a backbone gives our approach a boost across metrics
which is not surprising as larger models tend to have better
generation and instruction-following capabilities. Interestingly, the
fine-tuned model (FT-Llama 8B) trails behind our modular system when
using a backbone LLM of the same scale (Aspect-aware
decomposition-Llama 8B), both in terms of aspect coverage and opinion
faithfulness. Overall, our results suggest that prompt decomposition
is useful in opinion summarization and intermediate reasoning
steps based on task and domain-specific knowledge lead to meta-reviews
of higher quality.



\begin{figure}[t]
\centering
\includegraphics[width=0.49\textwidth, trim=0 0 0 0, clip]{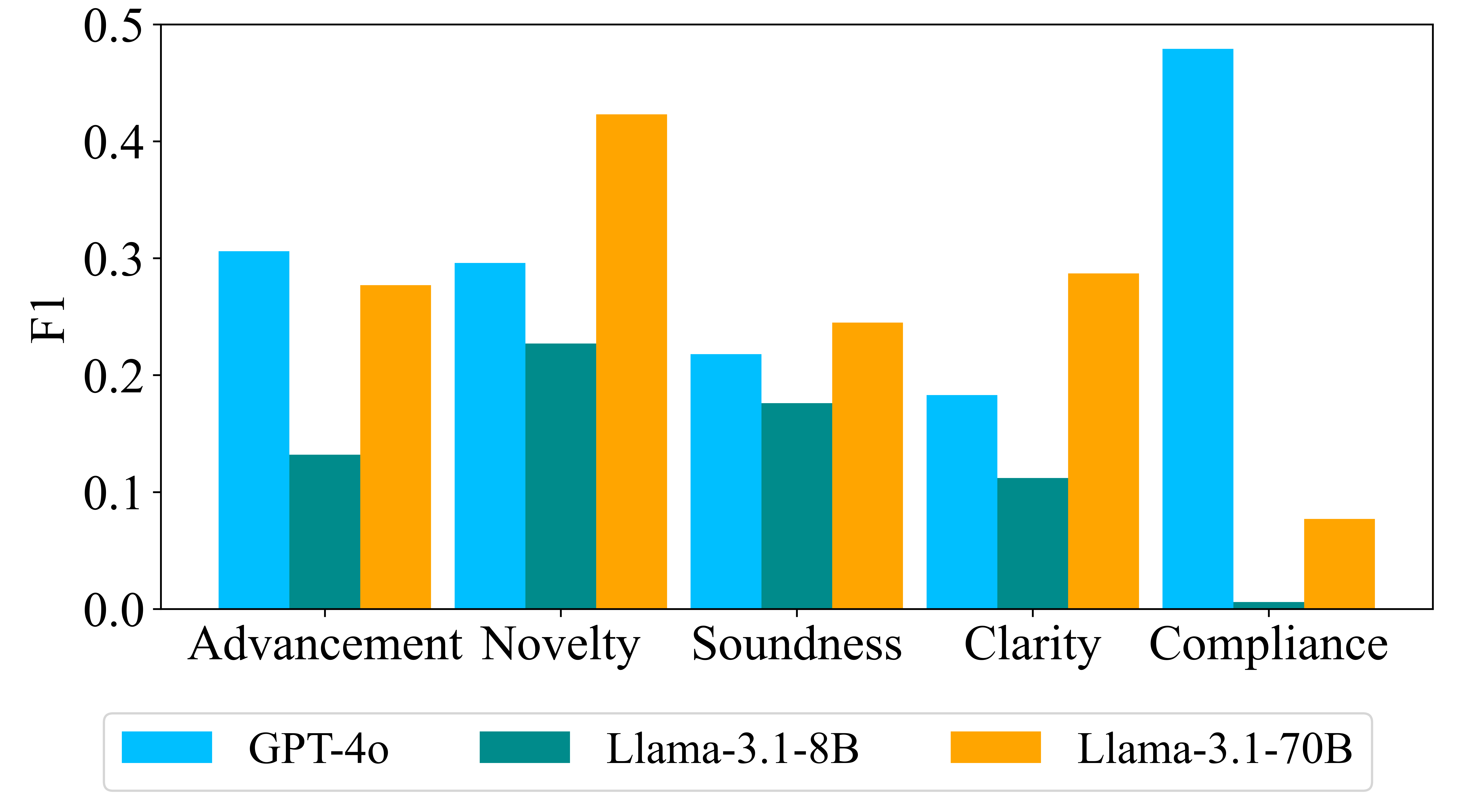}
\caption{Evaluation of text fragments extracted for individual review aspects by \textit{Aspect Identification}. }
\label{fig:grouping}
\end{figure}

\paragraph{Llama-70B performs well at identifying and summarizing
  aspects.}  In addition to evaluating the generated meta-reviews, we
conduct evaluations on the intermediate outputs of our modules. We
only report results on the scientific domain reusing the ground truth
annotations\footnote{\scriptsize
https://github.com/oaimli/MetaReviewingLogic} provided in
\citet{meta_review_logic_2024}. For \textit{Aspect Identification}, we
calculate word-level {Recall},
{Precision}, and F$_1$ between model-extracted text
fragments and human-annotated text fragments
following~\citet{meta_review_logic_2024}. The scores shown in
\tabref{intermediate_selection} denote how accurately our approach
extracts opinionated text from source reviews. We find that Llama-3.1-70B is
the best model for this module, even better than GPT-4o (in terms of
F$_{1}$). Moreover, \figref{grouping} shows that Llama-3.1-70B also
performs well on individual review aspects, especially frequent ones
including \textit{Novelty}, \textit{Soundness} and
\textit{Clarity}. For \textit{Opinion Consolidation},
\tabref{intermediate_reasoning} shows that Llama-3.1-70B performs
better than other models at generating aspect-specific
meta-reviews. Taken together, the evaluations on intermediate outputs
explain Llama-3.1-70B's superior performance at the end task.

\begin{table}[t]
\small
    \setlength\tabcolsep{4pt}
    \centering
    \begin{adjustbox}{max width=0.90\linewidth}
    \begin{tabular}{lccc}
    \toprule
    \multicolumn{1}{c}{\textbf{Models}} & \textbf{Recall}$\uparrow$ & \textbf{Precision}$\uparrow$ & \textbf{F}$_1\uparrow$ \\
    \midrule
    GPT-4o        & \textbf{0.82} & 0.27 & 0.40 \\
    Llama-3.1-8B  & 0.80 & 0.25 & 0.38\\
    Llama-3.1-70B & 0.74 & \textbf{0.34} &\textbf{0.46} \\
    \bottomrule
    \end{tabular}
    \end{adjustbox}
    \caption{Evaluation of text fragments extracted by \textit{Aspect
        Identification} against human annotations.}
    \label{tab:intermediate_selection}
\end{table}

\begin{table}[t]
 \small
    \setlength\tabcolsep{4pt}
    \centering
    \begin{adjustbox}{max width=0.90\linewidth}
    \begin{tabular}{lcc}
    \toprule
    \multicolumn{1}{c}{\textbf{Models}} & \textbf{AlignScore-S}$\uparrow$ & \textbf{Rouge}$\uparrow$ \\
    \midrule
    GPT-4o & 0.86 & 18.40\\
    Llama-3.1-8B & 0.82 & \textbf{18.24}\\
    Llama-3.1-70B & \textbf{0.87} & 16.93 \\
    \bottomrule
    \end{tabular}
    \end{adjustbox}
    \caption{Evaluation of aspect-specific meta-reviews,
      i.e.,~intermediate outputs of \textit{Opinion Consolidation}.}
    \label{tab:intermediate_reasoning}
    \vspace{-1em}
\end{table}

\begin{table}[t]
 \small
    \setlength\tabcolsep{3.5pt}
    \centering
    \begin{adjustbox}{max width=0.98\linewidth}
    \begin{tabular}{clcc}
    \toprule
    \textbf{Domain} & \textbf{Modules} & \textbf{Coverage}$\uparrow$ & \textbf{AlignScore-S}$\uparrow$ \\
    \midrule
    \multirow{3}{*}{Hotels} & AI+OC+MS & \textbf{0.99} & 0.80 \\
    & OC+MS & \textbf{0.99} & \textbf{0.83} \\
    & AI+MS & 0.55 & 0.62 \\
    \midrule
    \multirow{3}{*}{Shoes} & AI+OC+MS & \textbf{0.83} & \textbf{0.74} \\
    & OC+MS & 0.69 & 0.72 \\
    & AI+MS & 0.61 & 0.69 \\
    \midrule
     & AI+OC+MS & 0.97 & \textbf{0.79} \\
   \multirow{1}{*}{Research} & OC+MS & \textbf{0.98} & {0.78} \\
   \multirow{1}{*}{Articles} & AI+MS & 0.97 & 0.75 \\
    & AI$^\dagger$+OC+MS & 0.97 & 0.69 \\
    \bottomrule
    \end{tabular}
    \end{adjustbox}
    \caption{Ablations quantifying the contribution of different
      modules on three domains (hotels, shoes, research articles). AI:
      \textit{Aspect Identification}, OC: \textit{Opinion
        Consolidation}, MS: \textit{Meta-Review Synthesis},
      AI$^\dagger$: text fragments selected by humans. Results shown
      for \emph{Aspect-aware decomposition-Llama 70B}.}
    \label{tab:ablations}
    \vspace{-1em}
\end{table}

\begin{table}[t]
 \small
    \setlength\tabcolsep{1pt}
    \centering
    \begin{adjustbox}{max width=1.01\linewidth}
    \begin{tabular}{@{~~}l@{~}c@{~}c@{~}c@{}}
    \toprule
    \multicolumn{1}{c}{\textbf{Model}} & \multicolumn{1}{c}{\textbf{Cover}$\uparrow$} & \multicolumn{1}{c}{\textbf{Faith}$\uparrow$} & \multicolumn{1}{c}{\textbf{Overall}$\uparrow$} \\
    \midrule
    \midrule
    \multicolumn{4}{c}{Research Articles}\\
    \midrule
    Sentiment CoT-GPT-4o & \hspace*{.13cm}0\% & \hspace*{.13cm}0\% & \hspace*{.13cm}0\%\\
    Human-written  reference & {\color{darkpastelred}{80\%}} & {\color{darkpastelred}{80\%}} & {\color{darkpastelred}{80\%}}\\
    Automatic decomposition-Llama 70B & {\color{darkpastelred}{90\%}} & {\color{darkpastelred}{90\%}} & {\color{darkpastelred}{90\%}}\\
    Chunk-wise decomposition-Llama 70B& {\color{darkpastelred}{70\%}} & {\color{darkpastelred}{90\%}} & {\color{darkpastelred}{90\%}}\\
    Naive aspect-aware prompting-Llama 70B & \hspace*{.13cm}0\% & \hspace*{.13cm}0\% & \hspace*{.13cm}10\% \\
    Aspect-aware decomposition-GPT-4o & 10\% & {\color{darkpastelred}{50\%}} & {\color{darkpastelred}{50\%}}\\
    \midrule
    \midrule
    \multicolumn{4}{c}{Sports Shoes} \\
    \midrule
    HIRO-abs &  \hspace*{.2cm}\color{darkpastelred}{90\%}
    & \color{darkpastelred}{90\%} & \hspace*{.13cm}\color{darkpastelred}{90\%} \\
    Human-written reference & \hspace*{.2cm}{\color{darkpastelred}{90\%}} & {\color{darkpastelred}{90\%}} & \hspace*{.13cm}{\color{darkpastelred}{90\%}}\\
    Automatic decomposition-Llama 70B & {\color{darkpastelred}{100\%}} & {\color{darkpastelred}{90\%}} & {\color{darkpastelred}{100\%}}\\
    Chunk-wise decomposition-Llama 70B & \hspace*{.2cm}{\color{darkpastelred}{80\%}} & {\color{darkpastelred}{80\%}} & \hspace*{.13cm}{\color{darkpastelred}{70\%}}\\
    Naive aspect-aware prompting-Llama 70B & \hspace*{.2cm}20\% & 20\% & \hspace*{.13cm}40\%\\
    Aspect-aware decomposition-GPT-4o & \hspace*{.2cm}10\% & 20\% & \hspace*{.13cm}30\%\\
    \midrule
    \midrule
    \multicolumn{4}{c}{Hotels}\\
    \midrule
    HIRO-abs & {\color{darkpastelred}{80\%}} & {\color{darkpastelred}{100\%}} & {\color{darkpastelred}{100\%}} \\
    Human-written reference & 30\% & {\color{darkpastelred}{70\%}} & {\color{darkpastelred}{100\%}}\\
    Automatic decomposition-Llama 70B & {\color{darkpastelred}{90\%}} & {\color{darkpastelred}{100\%}} & {\color{darkpastelred}{100\%}} \\
    Chunk-wise decomposition-Llama 70B & {\color{darkpastelred}{50\%}} & {\color{darkpastelred}{60\%}} & {\color{darkpastelred}{80\%}} \\
    Naive aspect-aware prompting-Llama 70B & {\color{darkpastelred}{100\%}} & {\color{darkpastelred}{100\%}} & {\color{darkpastelred}{100\%}}\\
    Aspect-aware decomposition-GPT-4o & \hspace*{.13cm}0\% & \hspace*{.13cm}0\% & \hspace*{.13cm}10\% \\
    \bottomrule
    \end{tabular}
    \end{adjustbox}
    \caption{Proportion of times (\%) crowdworkers preferred our model
      (\emph{Aspect-aware decomposition-Llama 70B}) against depicted
      systems. We highlight in {\color{darkpastelred}{red}}
      comparisons where our model is chosen as better more than~50\%
      of the time (higher is better). For example, `90\%' means that
      crowdworkers prefer our system on 9 out of 10 entities. We take a
      majority vote to determine a single system preference.}
    \label{tab:analysis_meta_reviews}
    \vspace{-1.5em}
\end{table}

\paragraph{\emph{Opinion Consolidation} is the most important module.}
We further examine the contributions of individual modules to
meta-review generation. Specifically, we perform two ablations: (1)
remove \textit{Aspect Identification} and directly generate
aspect-specific meta-reviews based on original reviews and (2) remove
\textit{Opinion Consolidation} and directly generate final
meta-reviews based on text fragments from \textit{Aspect
  Identification}. We use Llama-3.1-70B as our backbone LLM because of
its superior performance in previous experiments. As we have ground
truth text fragments for scientific reviews
\cite{meta_review_logic_2024}, we include another experiment in this
domain where we replace the output of \textit{Aspect Identification}
with human-annotated text fragments. According to~\tabref{ablations},
both \textit{Aspect Identification} and \textit{Opinion Consolidation} are crucial to generating more faithful meta-reviews
and with higher aspect coverage, however \emph{Opinion Consolidation}
appears to be the most critical as its removal decreases performance across
domains (exception: coverage for research articles). We also observe that model-extracted text fragments are on
par with human-selected ones but more helpful to generating faithful
meta-reviews.


\paragraph{Humans prefer meta-reviews generated by our modular system to
  gold-standard references.} We conduct  human evaluation to verify
that our approach generates meta-reviews that reflect the review aspects of
the input and are overall coherent and faithful. We recruited
crowdworkers through Prolific\footnote{\scriptsize
https://www.prolific.com/}, selected to be L1 English speakers from
the US or UK, and compensated above the UK living wage at 12GBP/hr. We
ask crowdworkers to read a set of source reviews followed by two
generated meta-reviews and select which meta-review is best (allowing
for ties) along two dimensions, as well as an overall preference:
\begin{itemize}
  \itemsep0em 
\item \textbf{Coverage} --- Which meta-review covers more review aspects in the source reviews?
\item \textbf{Faithfulness} --- Which meta-review has a higher percentage of
  opinions supported by the source reviews?
  \item \textbf{Overall} --- Which meta-review do you think is better
    overall?
\end{itemize}

We randomly select ten entities for each dataset (SPACE, AmaSum, and PeerSum) and construct six pairwise
combinations between our approach (Aspect-aware decomposition with
Llama-3.1-70B) and the systems shown
in~\tabref{analysis_meta_reviews}, including human-written reference
meta-reviews. For AmaSum and SPACE, we only present crowdworkers
with~$20\%$ of the reviews for each entity, to maintain a reasonable
workload (reviews are sampled randomly).
We elicit three annotations for each pairwise combination of system
outputs, leading to a total of~1,260 ratings. 
Annotators have reasonable agreement, with average values of
Krippendorff's $\alpha$ being 0.335 on shoes, 0.622 on hotels, and
0.463 on research articles. More details on experimental design and the
full instructions are in~\appref{sec:details_human_meta}.

\tabref{analysis_meta_reviews} shows the proportion of times (\%)
crowdworkers prefer our approach against a comparison system.  We
find that human judgments are broadly consistent with automatic
evaluation. Crowdworkers prefer our system to human references on two
(shoes and research articles) out of three domains. We consistently win
against automatic and chunk-wise decompositions (with Llama 70B),
but lose against our own decompositions with GPT-4o.



\begin{table}[t]
     \small
    \setlength\tabcolsep{4pt}
    \centering
    \begin{adjustbox}{max width=0.98\linewidth}
    \begin{tabular}{lcc}
    \toprule
    \textbf{Present Reasoning Steps} & \textbf{Time}$\downarrow$ & \textbf{Preferred}$\uparrow$ \\
    \midrule
    No reasoning steps & 10.9 & 20\% \\
    Automatic decomposition & 10.3 & 20\%\\
    Aspect-aware decomposition (ours) & \hspace{.2cm}9.3 & 40\%\\
    \bottomrule
    \end{tabular}
    \end{adjustbox}
    \caption{Average time (in minutes) humans take to write scientific
      meta-reviews and the proportion of times participants prefer
      meta-reviews when present with different intermediate reasoning
      steps (in exhausted pair-wise comparison). 
        }
    \label{tab:analysis_rationales}
    \vspace{-1.5em}
\end{table}

\paragraph{Aspect-aware decomposition allows humans to create better
summaries faster.}  We also evaluate the intermediate outputs produced
by our modules. In particular, we examine whether the specific module
decomposition adopted by our system is useful for real-world
meta-review writing. We ask annotators to write meta-reviews for hotel
reviews in three conditions: (1)~they are not given any intermediate
reasoning steps; (2) they are
given reasoning steps produced by automatic knowledge-agnostic decomposition
from \emph{Automatic decomposition-Llama 70B}; and (3) they are provided with
the intermediate outputs of our modules with \emph{Aspect-aware decomposition-Llama 70B} as reasoning steps. We record the time it takes crowdworkers to finish the writing.

We randomly select ten entities and obtain three meta-reviews for each
(according to the three conditions described above). We recruit five
annotators, however, each annotator writes a meta-review for each
entity once to avoid memorization. Based on the time reported
in~\tabref{analysis_rationales}, we find that providing intermediate outputs of our aspect-aware decomposition accelerates participants' writing compared with the other two conditions, reducing the time of writing a meta-review by~14.7\% (on
average). More details about how we present different reasoning steps to annotators and annotation instructions are provided
in~\appref{sec:details_human_intermediate}. 

We also ask another set of
annotators to assess the meta-reviews written above, by presenting
pair-wise comparisons (following the instructions of human annotation presented
in the previous section). We find that participants prefer
meta-reviews written based on the outputs of our modules twice as much
compared to the other two settings (Krippendorff's $\alpha$ is 0.542).

\section{Conclusion}

We propose modular decomposition for opinion summarization based on
review aspects. Our decomposition is evidence-based (the output of
each module can be traced back to its input), enabling greater
transparency and ease of inspection.  Extensive experiments
demonstrate that our modular framework outperforms state-of-the-art
methods and other strong baselines in multiple domains.  Human
evaluations reveal that our approach not only produces higher-quality
meta-reviews but also generates more useful intermediate outputs to
assist humans in composing meta-reviews. While our work focuses on
opinion summarization, the concept of aspect-aware decomposition
holds promise for other  complex language generation
tasks.

\section*{Limitations}
Despite promising results, our experimental findings are currently limited to English. Furthermore, the prompts used in 
our modular approach were not  optimized in any way, suggesting a potential area for future improvement. 

Our approach operates on a set of predefined aspects, whose definitions were sourced from  previous work~\citep{space_2021, meta_review_logic_2024}. While reviewing platforms often feature similar criteria (e.g.,   \url{https://runrepeat.com/hoka-bondi-8} for the domain of sports shoes), our work prioritizes enhancing the grounding and transparency of opinion summarization. Therefore, the broader task of aspect engineering, including unsupervised methods for aspect discovery, falls outside the scope of this investigation. 

Our meta-review synthesis module, aims to cover all aspects mentioned in the original reviews without considering their relative importance. We could easily filter the aspects based on the size of their corresponding  text fragments. However, we found this approach is not universally applicable across all domains. In the scientific review domain, for example, \textit{Advancement} opinions are more frequent than \textit{Novelty} ones (25\% vs. 14\%) but both aspects are equally important~\citep{meta_review_logic_2024}. We defer further investigation into aspect importance to  future work.

Finally, our approach does not
explicitly address the potential generation of biased or harmful
content, even though our goal is to ensure that the generated
meta-reviews remain grounded in the original content.

\section*{Ethics Statement}
Our work primarily focuses on enhancing the capabilities of AI systems
to assist humans, rather than aiming to replace them. As demonstrated
in our experiments, the intermediate outputs generated by our approach
can help humans produce higher-quality meta-reviews with
greater efficiency.

\section*{Acknowledgments}
The authors would like to thank the anonymous reviewers for their valuable feedback, which greatly helped improve this work.
We would also like to thank Tom Hosking and Alex Gurung for their help with the human annotation experiments. 
%
This work was partially supported by the Australian Research Council Discovery Grant awarded to Jey Han Lau (ID: DP240101006) and the UK Engineering and Physical Sciences Research Council (grant EP/W002876/1) awarded to Mirella Lapata.



\bibliography{references}
\bibliographystyle{acl_natbib}

\newpage
\onecolumn
\appendix

\section{Prompts for Aspect-aware Decomposition}
\label{sec:appendix_prompts_modular_approach}

In this section we provide the prompt templates used to decompose
opinion summarization into the modules of \emph{Aspect
Identification}, \emph{Opinion Consolidation}, and \emph{Meta-review
Synthesis}. Domain-specific prompts are provided in
Sections~\ref{sec:prompts_scientific}--\ref{sec:prompts_product}. 

\begin{figure}[h]
\small
  \begin{tcolorbox}[colback=gray!10!white,colframe=WildStrawberry,title=Aspect
    Identification,fonttitle=\bfseries, halign title=flush
    center]

    {You are good at understanding documents with \{\textcolor{red}{domain}\} review opinions.}\\
    {Below is a \{\textcolor{red}{domain}\} review for an academic manuscript, please extract fragments that are related to \{\textcolor{red}{the-review-aspect\}} of the \{\textcolor{red}{the entity}\}.}\\
    {Definition of \{the review aspect\}:\{the definition of the review aspect\}}\\
    {Example input review:}\\
    \{\textcolor{red}{the example input review}\}\\
    {Example format of extracted fragments in different lines:}\\
    \{\textcolor{red}{the example output}\}\\
    {Target input review:}\\
    \{\textcolor{red}{input-document}\}\\
    {Final extracted fragments (follow the format above in different
      lines and if no resulted fragments just output "No related
      fragments"):}\\
    \vspace*{-.2cm}
\end{tcolorbox}
    \caption{The few-shot prompt template for the \emph{Aspect Identification} module;
      text fragments are extracted for each (domain) aspect. Please note that for research articles we use few-shot prompting to enable the model follow the output format while for sports shoes and hotels zero-shot prompting (with just removing the demonstration example) could get reasonable performances.}
    \label{tab:prompt_extraction}
\end{figure}

\begin{figure}[h]
\small
  \begin{tcolorbox}[colback=gray!10!white,colframe=darkpastelred,title=Opinion 
    Consolidation,fonttitle=\bfseries, halign title=flush
    center]
{You are good at writing summaries for opinionated texts. You are given some opinionated text fragments, please write a concise summary for them.}\\
{Example input review fragments:}\\
\{\textcolor{red}{the example text fragments}\}\\
{Example summary of the input fragments:}\\
{\{\textcolor{red}{the example aspect-specific meta-review of the input fragments}\}}\\
{Target input fragments:}\\
\{\textcolor{red}{input-fragments}\}\\
{The final summary of these target input text fragments (just output
  the answer without any other content):}\\
    \vspace*{-.2cm}
    \end{tcolorbox}
\caption{The few-shot prompt template for the \textit{Opinion Consolidation} module; it
  outputs summaries for  individual review aspects. Please note that for research articles we use few-shot prompting to get better performance while for sports shoes and hotels zero-shot prompting (with just removing the demonstration example) could get reasonable performances.}
    \label{tab:prompt_reasoning}
\end{figure}

\begin{figure}[th!]
\small
  \begin{tcolorbox}[colback=gray!10!white,colframe=brown,title=Meta-Review
    Synthesis,fonttitle=\bfseries, halign title=flush
    center]
    {You are good at understanding documents with \{\textcolor{red}{domain}\} review opinions.}\\
    {Below are comments on different review aspects for \{\textcolor{red}{the entity}\}, please write a concise and natural meta-review which summaries the provided comments and covers all mentioned review aspects.}\\
    {Comments on different aspects:}\\
    \{\textcolor{red}{meta-reviews of individual review aspects}\}\\
{The meta-review is (directly output the answer without any
      other content):}
\end{tcolorbox}
    \caption{The prompt template for the \textit{Meta-Review Synthesis} module
      based on aspect-specific meta-reviews from the \emph{Opinion Consolidation} module. As zero-shot prompting gives us reasonable performances on all the three datasets, we used the same zero-shot prompt template for the module.}
    \label{tab:prompt_generation}
\end{figure}


\section{Prompts for Scientific Reviews of Research Articles}
\label{sec:prompts_scientific}


Prompts for \textit{Aspect Identification} 
are given in
Tables~\ref{tab:prompt_selection_advancement}--\ref{tab:prompt_selection_novelty}
for the aspects  \textit{Advancement}, \textit{Clarity},
\textit{Compliance}, \textit{Soundness}, and \textit{Novelty}.
The prompt for \emph{Opinion Consolidation} is  in \tabref{prompt_reasoning_paper} and all aspects share the same prompt for this module.
  The prompt for \emph{Meta-Review Synthesis} is in \tabref{prompt_generation_paper}.

\begin{figure}[h!]
  \scriptsize
\begin{tcolorbox}[colback=gray!10!white,colframe=WildStrawberry,title=Aspect
    Identification: Advancement,fonttitle=\bfseries, halign title=flush
    center]
    {You are good at understanding documents with scientific review opinions.}\\
    {Below is a scientific review for an academic manuscript, please extract text fragments that are related to Advancement of the research work.}\\

    {Definition of Advancement:}\\

    {Importance of the manuscript to discipline, significance of the contributions of the manuscript, and its potential impact to the field.}\\

    {Example input review:}\\

    {This paper theoretically studied one of the fundamental issue in CycleGAN (recently gained much attention for image-to-image translation). The authors analyze the space of exact and approximated solutions under automorphisms. Reviewers mostly agree with theoretical value of the paper. Some concerns on practical values are also raised, e.g., limited or no-surprising experimental results. In overall, I think this is a boarderline paper. But, I am a bit toward acceptance as the theoretical contribution is solid, and potentially beneficial to many future works on unpaired image-to-image translation.}\\

    {Example output fragments in different lines:}\\

    {Some concerns on practical values are also raised, e.g., limited or no-surprising experimental results.}\\

    {Reviewers mostly agree with theoretical value of the paper.}\\

    {But, I am a bit toward acceptance as the theoretical contribution is solid, and potentially beneficial to many future works on unpaired image-to-image translation.}\\

    {Target input review:}\\

    {\{\{input\_document\}\}}

    {Final extracted fragments (follow the format above in different
      lines and if no resulted fragments just output "No related
      fragments"):}\\
    \vspace*{.5cm}
    \end{tcolorbox}
    \caption{The prompt of \emph{Aspect Identification} for the aspect \textit{Advancement}.}
    \label{tab:prompt_selection_advancement}
\end{figure}
\begin{figure}[h!]
 \scriptsize
\begin{tcolorbox}[colback=gray!10!white,colframe=WildStrawberry,title=Aspect
    Identification: Clarity,fonttitle=\bfseries, halign title=flush
    center]
    {You are good at understanding documents with scientific review opinions.}\\
    {Below is a scientific review for an academic manuscript, please extract fragments that are related to Clarity of the research work.}\\

    {Definition of Clarity:}\\

    {The readability of the writing (e.g., structure and language), reproducibility of details, and how accurately what the research question is, what was done and what was the conclusion are presented.}\\

    {Example input review:}\\

    {The paper is about a software library that allows for relatively easy simulation of molecular dynamics. The library is based on JAX and draws heavily from its benefits.}\\

    {To be honest, this is a difficult paper to evaluate for everyone involved in this discussion. The reason for this is that it is an unconventional paper (software) whose target application centered around molecular dynamics. While the package seems to be useful for this purpose (and some ML-related purposes), the paper does not expose which of the benefits come from JAX and which ones the authors added in JAX MD. It looks like that most of the benefits are built-in benefits in JAX. Furthermore, I am missing a detailed analysis of computation speed (the authors do mention this in the discussion below and in a sentence in the paper, but this insufficient). Currently, it seems that the package is relatively slow compared to existing alternatives.}\\

    {Here are some recommendations:}\\
    {1. It would be good if the authors focused more on ML-related problems in the paper, because this would also make sure that the package is not considered a specialized package that overfits to molecular dynamics.}\\
    {2. Please work out the contribution/delta of JAX MD compared to JAX.}\\
    {3. Provide a thorough analysis of the computation speed.}\\
    {4. Make a better case, why JAX MD should be the go-to method for practitioners.}\\

    {Overall, I recommend rejection of this paper. A potential re-submission venue could be JMLR, which has an explicit software track.}\\

    {Example output fragments in different lines:}\\

    {While the package seems to be useful for this purpose (and some ML-related purposes), the paper does not expose which of the benefits come from JAX and which ones the authors added in JAX MD.}\\

    {Make a better case, why JAX MD should be the go-to method for practitioners.}\\

    {Target input review:}\\

    {\{\{input\_document\}\}}\\

    {Final extracted fragments (follow the format above in different
      lines and if no resulted fragments just output "No related
      fragments"):}\\
    \vspace*{.5cm}
        \end{tcolorbox}

    \caption{The prompt of \textit{Aspect Identification} for the aspect of \textit{Clarity}.}
    \label{tab:prompt_selection_clarity}
\end{figure}

\begin{figure}[ht!]
 \scriptsize
\begin{tcolorbox}[colback=gray!10!white,colframe=WildStrawberry,title=Aspect
    Identification: Compliance,fonttitle=\bfseries, halign title=flush
    center]
    {You are good at understanding documents with scientific review opinions.}\\
    {Below is a scientific review for an academic manuscript, please extract fragments that are related to Compliance of the research work.}\\

    {Definition of Compliance:}\\

    {Whether the manuscript fits the venue, and all ethical and publication requirements are met.}\\

    {Example input review:}\\

    {"The paper proposes a method to identify and correct regions on the data manifold in which a trained classifier fails. The *identification* phase is based on clustering classification failure regions in a GAN latent space and the *correction* phase is based on fine-tuning the classifier with additional synthetic samples from the GAN. The proposed method is strongly based on Zhao et al 2018 (Generating Natural Adversarial Examples), a method to generate on-manifold black-box adversarial examples using a GAN. The authors of the current paper describe some differences of their identification step from Zhao et al (end of section 3.2.1), but in my opinion they are minor. The main contribution of the current paper over Zhao et al seems to be clustering the adversarial examples (using GMM) and using them to fine-tune the classifier. This, in my opinion, is potentially an interesting idea, however, the authors do not show sufficient evidence of its success. Specifically, the authors claim to "achieve near perfect failure scenario accuracy with minimal change in test set accuracy", but they do not provide any details (e.g. table of accuracy values on the train, test and adversarial sets before and after the fine-tuning). I would also expect to see an ablation study comparing the proposed method to simply including the adversarial examples found using Zhao et al (w/o GMM fitting and sampling) as additional training example - a standard adversarial defense approach (see e.g. [1]).Perhaps more importantly, the objective of the proposed method is not, in my opinion, clear. The title and abstract describe the goal as "debugging" a classifier and correcting fail regions, however the described method seems like a defense against on-manifold adversarial attack. If the method, as claimed, helps debugging and correcting the classifier, I would expect to see an improved accuracy on the (natural) unseen test set - not just on the synthetically generated adversarial examples. The quality and clarity of the writing can be improved as well. A lot of space is allocated to describing well-known methods (e.g. VAE, GMM), however, critical information about the experimental results are missing. I'm also not sure all the formally defined algorithms and equations actually help in the understanding (e.g. algorithm 1, equation 2). Some of the mathematical notations are not standard. Minor comment: The norm in definition 3.1 is a regular vector norm (l2?) and not a matrix norm. To summarize: pros: - interesting idea (clustering on-manifold failures, labeling them and then using them to improve the classifier)cons:- contribution over Zhao et al not well established- insufficient and inaccurate experimental results- general quality of writing - not sure actual work and experiments match the stated objective - significance *Update:* Following the authors' response, I upgraded my rating, but I still think there are critical issues with the paper. The most problematic point, in my opinion, is the only-marginal improvement on the test data, indicating that the suggested training method only improves the specific "failure scenarios", making it is similar to adversarial training methods used to gain adversarial robustness. However, the abstract and introduction indicates that the paper helps in debugging in fixing failures in general, which, I think should have been evident in improved test accuracy.[1] Zhang, Hongyang, et al. "Theoretically principled trade-off between robustness and accuracy."ICML 2019}\\

    {Example output fragments in different lines:}\\

    {Some of the mathematical notations are not standard.}\\

    {Target input meta-review:}\\

    {\{\{input\_document\}\}}\\

    {Final extracted fragments (follow the format above in different lines and if no resulted fragments just output "No related fragments"):}\\
    \vspace*{-.2cm}
    \end{tcolorbox}
\vspace*{-.2cm}
\caption{The prompt of \emph{Aspect Identification} for the aspect of \textit{Compliance}.}
    \label{tab:prompt_selection_compliance}
\end{figure}

\begin{figure}[h!]
 \scriptsize
\begin{tcolorbox}[colback=gray!10!white,colframe=WildStrawberry,title=Aspect
    Identification: Soundness,fonttitle=\bfseries, halign title=flush
    center]
    {You are good at understanding documents with scientific review opinions.}\\
    {Below is a scientific meta-review for an academic manuscript, please extract fragments that are related to Soundness of the research work.}\\

    {Definition of Soundness:}
    {There are usually two types of soundness: (1) Empirical: how well experiments are designed and executed to support the claims, whether methods used are appropriate, and how correctly the data and results are reported, analysed, and interpreted. (2) Theoretical: whether arguments or claims in the manuscript are well supported by theoretical analysis, i.e., completeness, and the methodology (e.g., mathematical approach) and the analysis is correct.}\\

    {Example input meta-review:}\\

    {The paper proposes to use the mirror descent algorithm for the binary network. It is easy to read. However, novelty over ProxQuant is somehow limited. The theoretical analysis is weak, in that there is no analysis on the convergence and neither how to choose the projection for mirror mapping construction. Experimental results can also be made more convincing, by adding comparisons with bigger datasets, STOA networks, and ablation study to demonstrate why mirror descent is better than proximal gradient descent in this application.}\\

    {Example output fragments in different lines:}\\

    {The theoretical analysis is weak, in that there is no analysis on the convergence and neither how to choose the projection for mirror mapping construction.}\\

    {Experimental results can also be made more convincing, by adding comparisons with bigger datasets, STOA networks, and ablation study to demonstrate why mirror descent is better than proximal gradient descent in this application.}\\

    {Target input meta-review:}\\

    {\{\{input\_document\}\}}\\

    {Final extracted fragments (follow the format above in different lines and if no resulted fragments just output "No related fragments"):}\\
\vspace*{-.2cm}
\end{tcolorbox}
\vspace*{-.2cm}
\caption{The prompt of \emph{Aspect Identification} for the aspect of \textit{Soundness}.}
    \label{tab:prompt_selection_soundness}
\end{figure}

\begin{figure}[ht!]
  \scriptsize
\begin{tcolorbox}[colback=gray!10!white,colframe=WildStrawberry,title=Aspect
    Identification: Novelty,fonttitle=\bfseries, halign title=flush
    center]
    {You are good at understanding documents with scientific review opinions.}\\
    {Below is a scientific meta-review for an academic manuscript, please extract fragments that are related to Novelty of the research work.}\\

    {Definition of Novelty:}\\

    {How original the idea (e.g., tasks, datasets, or methods) is, and how clear where the problems and methods sit with respect to existing literature (i.e., meaningful comparison).}\\

    {Example input meta-review:}\\

    {The manuscript describes a method for identifying and correcting classifier performance when labels are assigned incorrectly. The identification is based on clustering classification failure regions in a VAE latent space and the correction phase is based on fine-tuning the classifier with additional synthetic samples from the VAE.}\\

    {Reviewers agreed that the manuscript is not ready for publication. The main issue is that the suggested training method is similar to adversarial training methods used to gain adversarial robustness. The method does not help in debugging and fixing failures in general.}\\

    {Example output fragments in different lines:}\\

    {Reviewers appreciated the novelty, introducing a new simpler routing mechanism, and achieving good performance on real world datasets.}\\

    {In particular, removing the squash function and experimenting with concurrent routing was highlighted as significant progress.}\\

    {Alongside with them, I acknowledge the novelty of using layer norm and parallel execution, and recommend accept.}\\

    {Target input meta-review:}\\

    {\{\{input\_document\}\}}\\

    {Final extracted fragments (follow the format above in different lines and if no resulted fragments just output "No related fragments"):}\\
    \end{tcolorbox}
    \caption{The prompt of \emph{Aspect Identification} for the aspect of \emph{Novelty}.}
    \label{tab:prompt_selection_novelty}
\end{figure}

\begin{figure}[ht!]
  \scriptsize
\begin{tcolorbox}[colback=gray!10!white,colframe=darkpastelred,title=Opinion
    Consolidation,fonttitle=\bfseries, halign title=flush
    center]

    {You are good at writing summaries for opinionated texts. You are given some opinionated text fragments, please write a concise summary for them.}

    {Example input review fragments:}\\

    {"1) **Evaluating different explanation techniques:**",}\\
    {"We thus believe that our results do *not* violate the surmise made in the shared reference, but rather support it.",}\\
    {"We believe this makes our findings generalizable.",}\\
    {"Although, the paper brings out the importance of analogies as explanations (which further motivates our work)",}\\
    {"The proposed technique is flexible as it can provide two forms of explanations: feature and analogy-based.",}\\
    {"Moreover, explanations in the form of analogies are intuitive for human users.",}\\
    {"We feel that analogous examples do not need to share common words, content, or sentence structure. What is important is that they *point to latent factors* that may be responsible for the model's output.",}\\
    {"**Purpose of analogies:**",}\\
    ...\\
    {"The authors solved this problem by the use of a learned local distance matrix, in which interaction effects are clearly shown."}\\

    {Example summary of the input fragments:}\\

    {The proposed approach to explain similarity prediction is a relatively less explored area, which makes the problem addressed and the proposed method unique.}\\

    {Example input review fragments:}\\
    {"The paper is technically sound, and the claims are carefully developed and well supported.",}\\
    {"The manuscript is well structured and very clearly written, with helpful introductions to the methodological ingredients that it builds upon.",}\\
    {"The paper could be further improved with some reflection on the limitations of the approach.",}\\
    {"I am not certain how large a contribution it will have to the field of Bayesian inference in general.",}\\
    ...\\
    {"I'll use the rest of the section for high-level comments.",}\\
    {"- In its current form, the paper convinces me that SHF decreases runtime and increases performance for datasets with low complexity."}\\

    {Example summary of the input fragments:}\\

    {Based on these, I recommend acceptance for this paper. All reviewers agree that the paper proposes an interesting approach to Bayesian inference incorporating coresets with Hamiltonian flows.}\\

    {Target input review fragments:}\\

    {\{\{review\_fragments\}\}}\\

    {The final summary of these target input text fragments (just output the answer without any other content):}\\
\vspace*{-.3cm}
\end{tcolorbox}
\vspace*{-.3cm}
\caption{The prompt of \emph{Opinion Consolidation} for any aspect of scientific reviews.}
    \label{tab:prompt_reasoning_paper}
\end{figure}


\begin{figure}[ht!]
\scriptsize
  \begin{tcolorbox}[colback=gray!10!white,colframe=brown,title=Meta-review
    Synthesis,fonttitle=\bfseries, halign title=flush
    center]
    {You are good at understanding documents with scientific review opinions.}\\
    {Below are comments on different review aspects for an academic manuscript, please write a concise and natural meta-review which summaries the provided comments and covers all mentioned review aspects.}\\

    {Comments on different aspects:}\\

    {\{\{metas\_generated\}\}}\\

    {The meta-review is (directly output the answer without any other
      content):}\\
    \vspace{-.2cm}
 \end{tcolorbox}
    \caption{The prompt of \emph{Meta-Review Synthesis} for research articles.}
\vspace{-.2cm}
    \label{tab:prompt_generation_paper}
\end{figure}

\section{Prompts for  Business Reviews of Hotels}
\label{sec:prompts_business}

Prompts for \emph{Aspect Identification} on hotels are
shown in
Tables~\ref{tab:prompt_selection_building}--\ref{tab:prompt_selection_service}
for the aspects \emph{Building}, \emph{Cleanliness}, \emph{Food},
\emph{Location}, \emph{Rooms}, and \emph{Service}. The prompt for
     \emph{Opinon Consolidation} for any review aspect is in
     \tabref{prompt_reasoning_hotel}. The prompt for \emph{Meta-Review Synthesis} is present in \tabref{prompt_generation_hotel}.

     \begin{figure}[ht!]
       \scriptsize
\begin{tcolorbox}[colback=gray!10!white,colframe=WildStrawberry,title=Aspect
    Identification: Building,fonttitle=\bfseries, halign title=flush
    center]
    {You are good at understanding documents with hotel review opinions.}\\
    {Below is a business review for a hotel, please extract fragments that are related to Building of the hotel.}\\

    {Definition of Building:}\\
    {Analysis of how well the hotel was constructed, its design, functionality, and how these factors contribute to the success and satisfaction of its guests.}\\

    {Target input review:}\\

    {\{\{input\_document\}\}}\\

    {Final extracted fragments (follow the format above in different lines and if no resulted fragments just output "No related fragments"):}\\
\vspace{-.2cm}
\end{tcolorbox}
\vspace{-.2cm}
\caption{The prompt of \emph{Aspect Identification} for the aspect of \textit{Building}.}
    \label{tab:prompt_selection_building}
\end{figure}

     \begin{figure}[ht!]
       \scriptsize
\begin{tcolorbox}[colback=gray!10!white,colframe=WildStrawberry,title=Aspect
    Identification: Cleanliness,fonttitle=\bfseries, halign title=flush
    center]
    {You are good at understanding documents with hotel review opinions.}\\
    {Below is a business review for a hotel, please extract fragments that are related to Cleanliness of the hotel.}\\

    {Definition of Cleanliness:}\\
    {Evaluation of how well the hotel maintains a clean, sanitary, and comfortable environment for its guests, impacting their overall experience and satisfaction.}\\

    {Target input review:}\\

    {\{\{input\_document\}\}}\\

    {Final extracted fragments (follow the format above in different lines and if no resulted fragments just output "No related fragments"):}\\
\vspace{-.2cm}
\end{tcolorbox}
\vspace{-.2cm}
\caption{The prompt of \emph{Aspect Identification} for the aspect of \textit{Cleanliness}.}
    \label{tab:prompt_selection_cleanliness}
\end{figure}
     \begin{figure}[ht!]
       \scriptsize
\begin{tcolorbox}[colback=gray!10!white,colframe=WildStrawberry,title=Aspect
    Identification: Food,fonttitle=\bfseries, halign title=flush
    center]
    {You are good at understanding documents with hotel review opinions.}\\
    {Below is a business review for a hotel, please extract fragments that are related to Food of the hotel.}\\

    {Definition of Food:}\\
    {Evaluation of the dining experience including the quality and variety of the food, ultimately affecting guest satisfaction and the hotel’s reputation.}

    {Target input review:}

    {\{\{input\_document\}\}}\\

    {Final extracted fragments (follow the format above in different lines and if no resulted fragments just output "No related fragments"):}\\
\vspace{-.2cm}
\end{tcolorbox}
\vspace{-.2cm}
\caption{The prompt of \emph{Aspect Identification} with the aspect of \textit{Food}.}
    \label{tab:prompt_selection_food}
\end{figure}

     \begin{figure}[ht!]
       \scriptsize
\begin{tcolorbox}[colback=gray!10!white,colframe=WildStrawberry,title=Aspect
    Identification: Location,fonttitle=\bfseries, halign title=flush
    center]
    \vspace*{-.2cm}
    {You are good at understanding documents with hotel review opinions.}\\
    {Below is a business review for a hotel, please extract fragments that are related to Location of the hotel.}\\

    {Definition of Location:}\\
    {Analysis of how the hotel’s location influences the guest experience, considering factors like convenience, safety, proximity to attractions, and the overall environment.}\\

    {Target input review:}\\

    {\{\{input\_document\}\}}\\

    {Final extracted fragments (follow the format above in different lines and if no resulted fragments just output "No related fragments"):}\\
\vspace{-.3cm}
\end{tcolorbox}
\vspace{-.2cm}
    \caption{The prompt of \emph{Aspect Identification} for the aspect of \emph{Location}.}
    \label{tab:prompt_selection_location}
\end{figure}

\begin{figure}[ht!]
\scriptsize
  \begin{tcolorbox}[colback=gray!10!white,colframe=WildStrawberry,title=Aspect
    Identification: Rooms,fonttitle=\bfseries, halign title=flush
    center]
      \vspace*{-.2cm}
    {You are good at understanding documents with hotel review opinions.}\\
    {Below is a business review for a hotel, please extract fragments that are related to Rooms of the hotel.}\\

    {Definition of Rooms:}\\
    {Assessment of how well the room meets the guest’s needs and expectations in terms of comfort, cleanliness, amenities, and overall experience.}\\

    {Target input review:}\\

    {\{\{input\_document\}\}}\\

    {Final extracted fragments (follow the format above in different lines and if no resulted fragments just output "No related fragments"):}\\
\vspace{-.2cm}    
    \end{tcolorbox}
\vspace{-.2cm}    
  \caption{The prompt of \emph{Aspect Identification} for the review aspect of \textit{Rooms}.}
    \label{tab:prompt_selection_rooms}
\end{figure}

\begin{figure}[ht!]
  \scriptsize
\begin{tcolorbox}[colback=gray!10!white,colframe=WildStrawberry,title=Aspect
    Identification: Service,fonttitle=\bfseries, halign title=flush
    center]
    \vspace*{-.2cm}
{You are good at understanding documents with hotel review opinions.}\\
    {Below is a business review for a hotel, please extract fragments that are related to Service of the hotel.}\\

    {Definition of Service:}\\
    {Assessment of how well the hotel staff and management meet the needs of their guests, impacting their comfort, convenience, and overall experience.}\\

    {Target input review:}\\

    {\{\{input\_document\}\}}\\

    {Final extracted fragments (follow the format above in different lines and if no resulted fragments just output "No related fragments"):}\\
    
\vspace{-.2cm}    
    \end{tcolorbox}
\vspace{-.2cm}    
\caption{The prompt of \emph{Aspect Identification} with the aspect of \textit{Service}.}
    \label{tab:prompt_selection_service}
\end{figure}


\begin{figure}[ht!]
  \scriptsize
\begin{tcolorbox}[colback=gray!10!white,colframe=darkpastelred,title=Opinion
    Consolidation,fonttitle=\bfseries, halign title=flush
    center]
    \vspace*{-.2cm}
    {You are good at writing summaries for opinionated texts. You are given some opinionated text fragments, please write a concise summary for them.}\\

    {Target input review fragments:}\\

    {\{\{review\_fragments\}\}}\\

    {The final summary of these target input text fragments (just produce the answer without any other content):}\\
\vspace*{-.2cm}
\end{tcolorbox}
\vspace*{-.2cm}
    \caption{The prompt of \emph{Opinion Consolidation} for any individual review aspect for hotels.}
    \label{tab:prompt_reasoning_hotel}
\end{figure}


\begin{figure}[ht!]
  \scriptsize
\begin{tcolorbox}[colback=gray!10!white,colframe=brown,title=Meta-Review
    Synthesis,fonttitle=\bfseries, halign title=flush
    center]
  \vspace*{-.2cm}
    {You are good at understanding documents with hotel review opinions.}\\
    {Below are business reviews in different aspects for a hotel, please write a concise and natural meta-review which summaries the provided comments and covers all mentioned review aspects.}\\

    {Comments on different aspects:}\\

    {\{\{metas\_generated\}\}}

    {The meta-review is (directly output the answer without any other content):}\\
\vspace{-.2cm}    
\end{tcolorbox}
\vspace{-.2cm}    
\caption{The prompt of \emph{Meta-Review Synthesis} for hotels.}
    \label{tab:prompt_generation_hotel}
\end{figure}

\section{Prompts for Product Reviews of Sports Shoes}
\label{sec:prompts_product}

Prompts for \emph{Aspect Identification} are given in
Tables~\ref{tab:prompt_selection_breathability}--\ref{tab:prompt_selection_weight}
for the aspects \emph{Breathability}, \emph{Comfort},
\emph{Cushioning}, \emph{Durability}, \emph{Flexibility}, \emph{Misc},
\emph{Size and Fit}, \emph{Stability}, \emph{Traction}, and
\emph{Weight}.  The prompt for \emph{Opinion Consolidation} for any
 aspect is in \tabref{prompt_reasoning_shoes}. The prompt for \emph{Meta-Review Synthesis} is  in \tabref{prompt_generation_shoes}.

\begin{figure}[ht!]
\scriptsize
  \begin{tcolorbox}[colback=gray!10!white,colframe=WildStrawberry,title=Aspect
    Identification: Breathability,fonttitle=\bfseries, halign title=flush
    center]
        \vspace*{-.2cm}
    {You are good at understanding documents with sports shoes review opinions.}\\
    {Below is a product review for a pair of shoes, please extract fragments that are related to Breathability of shoes.}\\

    {Definition of Breathability:}\\
    {Evaluation about breathability of the shoes.}\\

    {Target input review:}\\

    {\{\{input\_document\}\}}\\

    {Final extracted fragments (follow the format above in different lines and if no resulted fragments just output "No related fragments"):}\\
\vspace*{-.2cm}
  \end{tcolorbox}
\vspace*{-.2cm}
  \caption{The prompt of \emph{Aspect Identification} for the aspect of \textit{Breathability}.}
    \label{tab:prompt_selection_breathability}
\end{figure}

\begin{figure}[ht!]
\scriptsize
  \begin{tcolorbox}[colback=gray!10!white,colframe=WildStrawberry,title=Aspect
    Identification: Comfort,fonttitle=\bfseries, halign title=flush
    center]
    \vspace*{-.2cm}
    {You are good at understanding documents with sports shoes review opinions.}\\
    {Below is a product review for a pair of shoes, please extract fragments that are related to Comfort of shoes.}\\

    {Definition of Comfort:}\\
    {Evaluation about comfort of the shoes, such as tongue padding, heel tab, and removable insole.}\\

    {Target input review:}

    {\{\{input\_document\}\}}\\

    {Final extracted fragments (follow the format above in different lines and if no resulted fragments just output "No related fragments"):}\\
\vspace*{-.2cm}
  \end{tcolorbox}
  \vspace*{-.2cm}
    \caption{The prompt of \emph{Aspect Identification} with the aspect of \textit{Comfort}.}
    \label{tab:prompt_selection_comfort}
\end{figure}

\begin{figure}[ht!]
\scriptsize
  \begin{tcolorbox}[colback=gray!10!white,colframe=WildStrawberry,title=Aspect
    Identification: Cushioning,fonttitle=\bfseries, halign title=flush
    center]
    \vspace*{-.2cm}
    {You are good at understanding documents with sports shoes review opinions.}\\
    {Below is a product review for a pair of shoes, please extract fragments that are related to Cushioning of shoes.}\\

    {Definition of Cushioning:}\\
    {Evaluation about cushioning of the shoes, such as heel stack and forefoot stack.}\\

    {Target input review:}\\

    {\{\{input\_document\}\}}\\

    {Final extracted fragments (follow the format above in different lines and if no resulted fragments just output "No related fragments"):}\\
\vspace*{-.2cm}
  \end{tcolorbox}
\vspace*{-.2cm}
  \caption{The prompt of \emph{Aspect Identification} for the review aspect of \emph{Cushioning}.}
    \label{tab:prompt_selection_cushioning}
\end{figure}

\begin{figure}[ht!]
\scriptsize
  \begin{tcolorbox}[colback=gray!10!white,colframe=WildStrawberry,title=Aspect
    Identification: Breathability,fonttitle=\bfseries, halign title=flush
    center]
    \vspace*{-.2cm}
{You are good at understanding documents with sports shoes review opinions.}\\
    {Below is a product review for a pair of shoes, please extract fragments that are related to Durability of shoes.}\\

    {Definition of Durability:}\\
    {Evaluation about durability of the shoes, such as outsole hardness and thickness.}\\

    {Target input review:}\\

    {\{\{input\_document\}\}}\\

    {Final extracted fragments (follow the format above in different lines and if no resulted fragments just output "No related fragments"):}\\
\vspace*{-.3cm}
  \end{tcolorbox}
\vspace*{-.2cm}
  \caption{The prompt of \emph{Aspect Identification} with the aspect of \textit{Durability}.}
    \label{tab:prompt_selection_durability}
\end{figure}

\begin{figure}[ht!]
\scriptsize
  \begin{tcolorbox}[colback=gray!10!white,colframe=WildStrawberry,title=Aspect
    Identification: Flexibility,fonttitle=\bfseries, halign title=flush
    center]

    {You are good at understanding documents with sports shoes review opinions.}\\
    {Below is a product review for a pair of shoes, please extract fragments that are related to Flexibility of shoes.}\\

    {Definition of Flexibility:}\\
    {Evaluation about flexibility of the shoes, such as stiffness, stiffness in the cold, and difference in stiffness in the cold.}\\

    {Target input review:}\\

    {\{\{input\_document\}\}}\\
    {Final extracted fragments (follow the format above in different lines and if no resulted fragments just output "No related fragments"):}\\
    \vspace*{-.2cm}
    \end{tcolorbox}
    \caption{The prompt of \emph{Aspect Identification} with the review aspect of \textit{Flexibility}.}
    \label{tab:prompt_selection_flexibility}
\end{figure}

\begin{figure}[ht!]
\scriptsize
  \begin{tcolorbox}[colback=gray!10!white,colframe=WildStrawberry,title=Aspect
    Identification: Misc,fonttitle=\bfseries, halign title=flush
    center]
    {You are good at understanding documents with sports shoes review opinions.}\\
    {Below is a product review for a pair of shoes, please extract fragments that are related to Misc of shoes.}\\

    {Definition of Misc:}\\
    {Evaluation about reflective elements of the shoes.}\\

    {Target input review:}\\

    {\{\{input\_document\}\}}\\

    {Final extracted fragments (follow the format above in different lines and if no resulted fragments just output "No related fragments"):}\\
\vspace*{-.2cm}
  \end{tcolorbox}
    \caption{The prompt of {Aspect Identification} with the review aspect of {Misc}.}
    \label{tab:prompt_selection_misc}
\end{figure}

\begin{figure}[ht!]
  \scriptsize
\begin{tcolorbox}[colback=gray!10!white,colframe=WildStrawberry,title=Aspect
    Identification: Size and Fit,fonttitle=\bfseries, halign title=flush
    center]

    {You are good at understanding documents with sports shoes review opinions.}\\
    {Below is a product review for a pair of shoes, please extract fragments that are related to Size and Fit of shoes.}\\

    {Definition of Size and Fit:}\\
    {Evaluation about size and fit of the shoes, such as internal length, toebox width at the widest part, and gusset type.}\\

    {Target input review:}\\

    {\{\{input\_document\}\}}\\

    {Final extracted fragments (follow the format above in different lines and if no resulted fragments just output "No related fragments"):}\\
    \vspace*{-.2cm}
    \end{tcolorbox}
    \caption{The prompt of \emph{Aspect Identification} for the aspect of \emph{Size and Fit}.}
    \label{tab:prompt_selection_sizefit}
\end{figure}

\begin{figure}[ht!]
\scriptsize
  \begin{tcolorbox}[colback=gray!10!white,colframe=WildStrawberry,title=Aspect
    Identification: Stability,fonttitle=\bfseries, halign title=flush
    center]
    {You are good at understanding documents with sports shoes review opinions.}\\
    {Below is a product review for a pair of shoes, please extract fragments that are related to Stability of shoes.}\\

    {Definition of Stability:}\\
    {Evaluation about stability of the shoes, such as torsional rigidity, heel counter stiffness, midsole width in the forefoot and midsole width in the heel.}\\

    {Target input review:}\\

    {\{\{input\_document\}\}}\\

    {Final extracted fragments (follow the format above in different lines and if no resulted fragments just output "No related fragments"):}\\
    
    \end{tcolorbox}
    \caption{The prompt of \emph{Aspect Identification} for the aspect of \emph{Stability}.}
    \label{tab:prompt_selection_stability}
\end{figure}

\begin{figure}[ht!]
\scriptsize
  \begin{tcolorbox}[colback=gray!10!white,colframe=WildStrawberry,title=Aspect
    Identification: Traction,fonttitle=\bfseries, halign title=flush
    center]
{You are good at understanding documents with sports shoes review opinions.}\\
    {Below is a product review for a pair of shoes, please extract fragments that are related to Traction of shoes.}\\

    {Definition of Traction:}\\
    {Evaluation about traction of the shoes, such as lug depth.}\\

    {Target input review:}\\

    {\{\{input\_document\}\}}\\

    {Final extracted fragments (follow the format above in different lines and if no resulted fragments just output "No related fragments"):}\\
    \end{tcolorbox}
    \caption{The prompt of \emph{Aspect Identification} for the review aspect of \textit{Traction}.}
    \label{tab:prompt_selection_traction}
\end{figure}

\begin{figure}[ht!]
\scriptsize
  \begin{tcolorbox}[colback=gray!10!white,colframe=WildStrawberry,title=Aspect
    Identification: Weight,fonttitle=\bfseries, halign title=flush
    center]
    {You are good at understanding documents with sports shoes review opinions.}\\
    {Below is a product review for a pair of shoes, please extract fragments that are related to Weight of shoes.}\\

    {Definition of Weight:}\\
    {Evaluation about weight of the shoes.}\\

    {Target input review:}\\

    {\{\{input\_document\}\}}\\

    {Final extracted fragments (follow the format above in different lines and if no resulted fragments just output "No related fragments"):}\\
    \end{tcolorbox}
    \caption{The prompt of \emph{Aspect Identification} for the review aspect of \textit{Weight}.}
    \label{tab:prompt_selection_weight}
\end{figure}

\begin{figure}[ht!]
\scriptsize
  \begin{tcolorbox}[colback=gray!10!white,colframe=darkpastelred,title=Opinion
    Consolidation,fonttitle=\bfseries, halign title=flush
    center]
    {You are good at writing summaries for opinionated texts. You are given some opinionated text fragments, please write a concise summary for them.}

    {Target input review fragments:}\\

    {\{\{review\_fragments\}\}}\\

    {The final summary of these target input text fragments (just produce the answer without any other content):}\\
    \end{tcolorbox}
    \caption{The prompt of \emph{Opinion Consolidation} for any individual review aspect for sports shoes.}
    \label{tab:prompt_reasoning_shoes}
\end{figure}


\begin{figure}[ht!]
\scriptsize
  \begin{tcolorbox}[colback=gray!10!white,colframe=brown,title=Meta-Review
    Synthesis,fonttitle=\bfseries, halign title=flush
    center]
    {You are good at understanding documents with sports shoes review opinions.}\\
    {Below are product reviews in different aspects for a pair of shoes, please write a concise and natural meta-review which summaries the provided comments and covers all mentioned review aspects.}\\

    {Comments on different aspects:}\\

    {\{\{metas\_generated\}\}}\\

    {The meta-review is (directly output the answer without any other content):}\\
\vspace*{-.2cm}
  \end{tcolorbox}
    \caption{The prompt of \emph{Meta-Review Synthesis} for the product reviews of sports shoes.}
    \label{tab:prompt_generation_shoes}
\end{figure}




\section{Implementation Details of Comparison Models}
\label{sec:implementation_baselines}

In this section we provide implementation details for the various
comparison models used in our experiments.
\begin{itemize}
    \item For HIRO-abs~\citep{hiro_2024}, we obtain generations for
    AmaSum and SPACE from https://github.com/tomhosking/hiro. There are three outputs for each entity and we use the first one as the generation of HIRO-abs.

    \item For TCG~\citep{prompted_opinion_summarization_2023}, we made some adaptation to get fair comparison. TCG only generates aspect-oriented summaries instead of an overall global summary, which we have to aggregate to ensure a fair comparison with our approach. We obtain their released aspect-oriented summaries and use the open-source Llama 70B to generate an overall summary. We use the same version of Llama 70B as in our experiments since the GPT-3.5 model used in their implementation has been deprecated.

    \item For fine-tuning Llama-3.1-8B, we trained the model with Transformers from Huggingface on the three datasets for 5 epochs on four NVIDIA A100 80G GPUs, with \texttt{{max-predict-length}=512}, \texttt{{bf16}=True}, \texttt{{batch-size}=1}, \texttt{{optim}=adafactor}, \texttt{{learning-rate}=1e-6},
  \texttt{{warmup-rate}=0.2}, \texttt{{label-smoothing-factor}=0.1},
  \texttt{{lr-scheduler-type}=cosine}, \texttt{{fsdp}=`full\_shard
    auto\_wrap offload'}.
 
    \item For \emph{naive aspect-aware prompting}, we only incorporate aspect
  descriptions into the prompt. As an example, we show the prompt for
  scientific reviews in~Figure~\ref{tab:prompt_knowledge_aware_prompting}.

    \item For \emph{Automatic decomposition}~\citep{decomposed_prompting_2023}, the prompting
  approach cannot be directly transferred to opinion
  summarization. Based on the idea of automatic decomposition, we
  implement automatic knowledge-agnostic decomposition on our experimental
  datasets. The idea is to first generate intermediate reasoning steps and then follow those steps in sequence to generate the final meta-review. We provide
example prompts for scientific reviews in Figure~\ref{tab:prompt_knowledge_agnostic_decomposition_steps} and~\ref{tab:prompt_knowledge_agnostic_decomposition_following}.

    \item For \emph{chunk-wise decomposition}~\citep{decomposed_prompting_2023},
  we first  generate small meta-reviews for each chunk, and then
  combine all chunk-specific meta-reviews with another prompt to generate the
  global meta-review. Example prompts for scientific reviews are
  shown in Figures\ref{tab:prompt_chunk_wise_decomposition_chunk} and
  \ref{tab:prompt_chunk_wise_decomposition_aggregation}.

\end{itemize}

\begin{figure}[t]
\small
  \begin{tcolorbox}[colback=gray!10!white,colframe=blue!10!gray,title=Naive Aspect-Aware
      Prompt,fonttitle=\bfseries, halign title=flush
    center]
    {Please write a summary for the reviews on a scientific article, focused on the review aspects below.}\\

    {Review aspects:}\\

    {(1) Advancement: importance of the manuscript to discipline, significance of the contributions of the manuscript, and its potential impact to the field.}\\

    {(2) Clarity: the readability of the writing (e.g., structure and language), reproducibility of details, and how accurately what the research question is, what was done and what was the conclusion are presented.}\\

    {(3) Compliance: whether the manuscript fits the venue, and all ethical and publication requirements are met.}\\

    {(4) Soundness: there are usually two types of soundness, empirical (how well experiments are designed and executed to support the claims, whether methods used are appropriate, and how correctly the data and results are reported, analysed, and interpreted.) and theoretical (whether arguments or claims in the manuscript are well supported by theoretical analysis, i.e., completeness, and the methodology, e.g., mathematical approach and the analysis is correct.)}\\

    {(5) Novelty: how original the idea (e.g., tasks, datasets, or methods) is, and how clear where the problems and methods sit with respect to existing literature (i.e., meaningful comparison).}\\

    {Reviews on a scientific article:}\\
    {\{\{source\_documents\}\}}\\

    {The output summary:}\\
\end{tcolorbox}
    \caption{The prompt with aspects in scientific reviews of research articles for \emph{naive aspect-aware prompting}.}
    \label{tab:prompt_knowledge_aware_prompting}
  \end{figure}

\begin{figure}[t]
\small
  \begin{tcolorbox}[colback=gray!10!white,colframe=blue!10!gray,title=Automatic Decomposition Prompt,fonttitle=\bfseries, halign title=flush
    center]
    {You are requested to write the steps. Please output the final answer with only the steps in different lines, no other useless content.}\\

    {Please give me sequential steps to write a summary specific for the following reviews on an academic paper.}\\
    {Reviews on a paper: \{source\_text\}}\\
    {The steps to write a summary in different lines:}\\
    
    \end{tcolorbox}
    \caption{The prompt for automatic decomposition to generate intermediate reasoning steps to write the meta-review for scientific reviews.}
    \label{tab:prompt_knowledge_agnostic_decomposition_steps}
\end{figure}

\begin{figure}[ht!]
\small
  \begin{tcolorbox}[colback=gray!10!white,colframe=blue!10!gray,title=
      Prompt to Follow Reasoning Steps from Automatic Decomposition,fonttitle=\bfseries, halign title=flush
    center]
    {You are requested to follow the instruction and only generate the requested output.}\\

    {\{output\_from\_last\_step\}}\\
    {Please follow the instruction below and give your output.}\\
    {\{current\_step\}}\\
    {The output:}\\
\vspace*{-.3cm}

  \end{tcolorbox}
  \caption{The prompt to follow automatically predicted steps by \emph{automatic decomposition} to generate the final meta-review.}
    \label{tab:prompt_knowledge_agnostic_decomposition_following}
\end{figure}

\begin{figure}[ht!]
\small
  \begin{tcolorbox}[colback=gray!10!white,colframe=blue!10!gray,title=Chunk
      Summarization Prompt,fonttitle=\bfseries, halign title=flush
    center]
    {You are requested to do summarization. Please output the final answer with only the summary, no other useless content.}\\

    {Please write a summary for the following review on an academic paper.}\\
    {The review: \{the\_text\_chunk\}}\\
    {The output summary:}\\
\vspace*{-.3cm}

  \end{tcolorbox}
    \caption{The prompt of \emph{chunk-wise decomposition} to summarize individual chunks of texts for scientific reviews of research articles.}
    \label{tab:prompt_chunk_wise_decomposition_chunk}
\end{figure}

\begin{figure}[ht!]
\small
  \begin{tcolorbox}[colback=gray!10!white,colframe=blue!10!gray,title=Summary
      Aggregation Prompt,fonttitle=\bfseries, halign title=flush
    center]
    {You are requested to do summarization. Please output the final answer with only the summary, no other useless content.}\\

    {Please write a summary for the following texts.}\\
    {The texts to be summarized:}\\
    {\{the\_concatenation\_of\_small\_meta\_reviews\_of\_chunks\}}\\
    {The output summary:}\\
    \vspace*{-.2cm}
    \end{tcolorbox}
    \caption{The Prompt for aggregating  chunk-specific meta-reviews into
      the global meta-review.}
    \label{tab:prompt_chunk_wise_decomposition_aggregation}
\end{figure}

\section{Implementation Details for Automatic Evaluation}
\label{sec:implementation_metrics}

Implementation details of G-Eval~\citep{geval_2021} are presented in
Figures~\ref{tab:prompt_geval_shoes}, \ref{tab:prompt_geval_papers},
and \ref{tab:prompt_geval_hotel} for the three domains,
respectively. We use gpt-4o-2024-05-13 as the backbone LLM of G-Eval.

\begin{figure}[ht!]
\small
  \begin{tcolorbox}[colback=gray!10!white,colframe=red!20!gray,title=G-Eval
      for Sports Shoes,fonttitle=\bfseries, halign title=flush
    center]
    {Here are several review documents that contain opinions from different people about a pair of shoes, along with a candidate summary of these reviews.}\\

    {You are required to evaluate how accurately the given summary reflects the overall opinions for review aspects expressed in the original reviews.}\\

    {Please read all opinions in the summary and calculate the percentage of faithful opinions that are clearly supported by the source review documents.}\\

    {Review documents:}\\

    {\{\{source\_documents\}\}}\\

    {The candidate summary:}\\

    {\{\{generation\_summary\}\}}\\

    {The percentage of faithful opinions (only output a decimal like 0.12, no other content):}\\
    \end{tcolorbox}
    \caption{The G-Eval prompt for evaluating meta-reviews for sports shoes.}
    \label{tab:prompt_geval_shoes}
\end{figure}

\begin{figure}[ht!]
\small
  \begin{tcolorbox}[colback=gray!10!white,colframe=red!20!gray,title=G-Eval
      for Research Articles,fonttitle=\bfseries, halign title=flush
    center]
    
    {Here are several review documents that contain opinions from different people about a scientific paper, along with a candidate summary of these reviews.}\\

    {You are required to evaluate how accurately the given summary reflects the overall opinions for review aspects expressed in the original reviews.}\\

    {Please read all opinions in the summary and calculate the percentage of faithful opinions that are clearly supported by the source review documents.}\\

    {Review documents:}\\

    {\{\{source\_documents\}\}}\\

    {The candidate summary:}\\

    {\{\{generation\_summary\}\}}\\

    {The percentage of faithful opinions (only output a decimal like 0.12, no other content):}\\
        \end{tcolorbox}
    \caption{The G-Eval prompt for evaluating  meta-reviews on research articles.}
    \label{tab:prompt_geval_papers}
\end{figure}

\begin{figure}[ht!]
\small
  \begin{tcolorbox}[colback=gray!10!white,colframe=red!20!gray,title=G-Eval
      for Hotels,fonttitle=\bfseries, halign title=flush
    center]
    
    {Here are several review documents that contain opinions from different people about a hotel, along with a candidate summary of these reviews.}\\

    {You are required to evaluate how accurately the given summary reflects the overall opinions for review aspects expressed in the original reviews.}\\

    {Please read all opinions in the summary and calculate the percentage of faithful opinions that are clearly supported by the source review documents.}\\

    {Review documents:}\\

    {\{\{source\_documents\}\}}\\

    {The candidate summary:}\\

    {\{\{generation\_summary\}\}}\\

    {The percentage of faithful opinions (only output a decimal like 0.12, no other content):}\\
    \end{tcolorbox}
    \caption{The G-Eval prompt for evaluating meta-reviews on hotels.}
    \label{tab:prompt_geval_hotel}
\end{figure}

\section{Details of Human Evaluation on Quality of Generated Meta-Reviews}
\label{sec:details_human_meta}

We conduct human evaluation based on pair-wise comparisons to verify
the quality of our generated meta-reviews (in terms of aspect coverage
and opinion faithfulness). We recruited crowdworkers through
Prolific\footnote{www.prolific.com} with compensation above the UK
living wage at \pounds12 per working hour.

For product reviews of sports shoes, we randomly select ten entities from the
test data of AmaSum. Based on generated meta-reviews, for each entity
we construct six pairs of comparisons between our modular approach
with Llama-3.1-70B as a backbone and comparison baselines. There are
originally about 400 source reviews in each entity and it is hard for humans
to review all of them. To balance annotator workload, we present
annotators with~$20\%$ reviews and randomly select reviews for three
times to ensure experimental consistency. 
Therefore, there are 18 pairs of comparisons for each
entity. Each pair is rated by three different annotators and we
obtain~540 annotations for the dataset.

\begin{figure*}[t]
\centering
\includegraphics[width=0.98\textwidth, trim=70 230 70 70, clip]{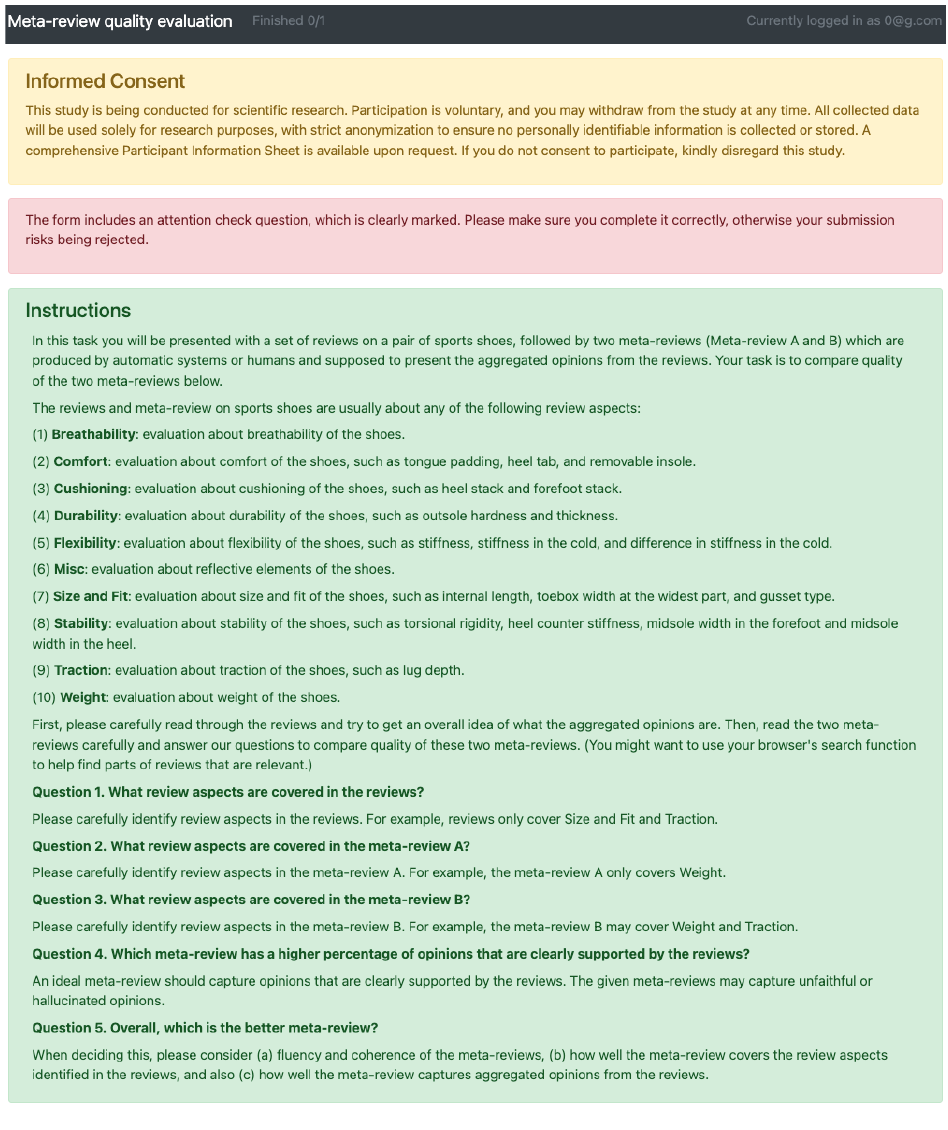}
\caption{Experimental instructions and interface for human evaluation
  study on   sports shoes reviews (part 1).}
\label{fig:meta_interface_shoes_p1}
\end{figure*}

\begin{figure*}[t]
\centering
\includegraphics[width=0.98\textwidth, trim=70 210 70 70, clip]{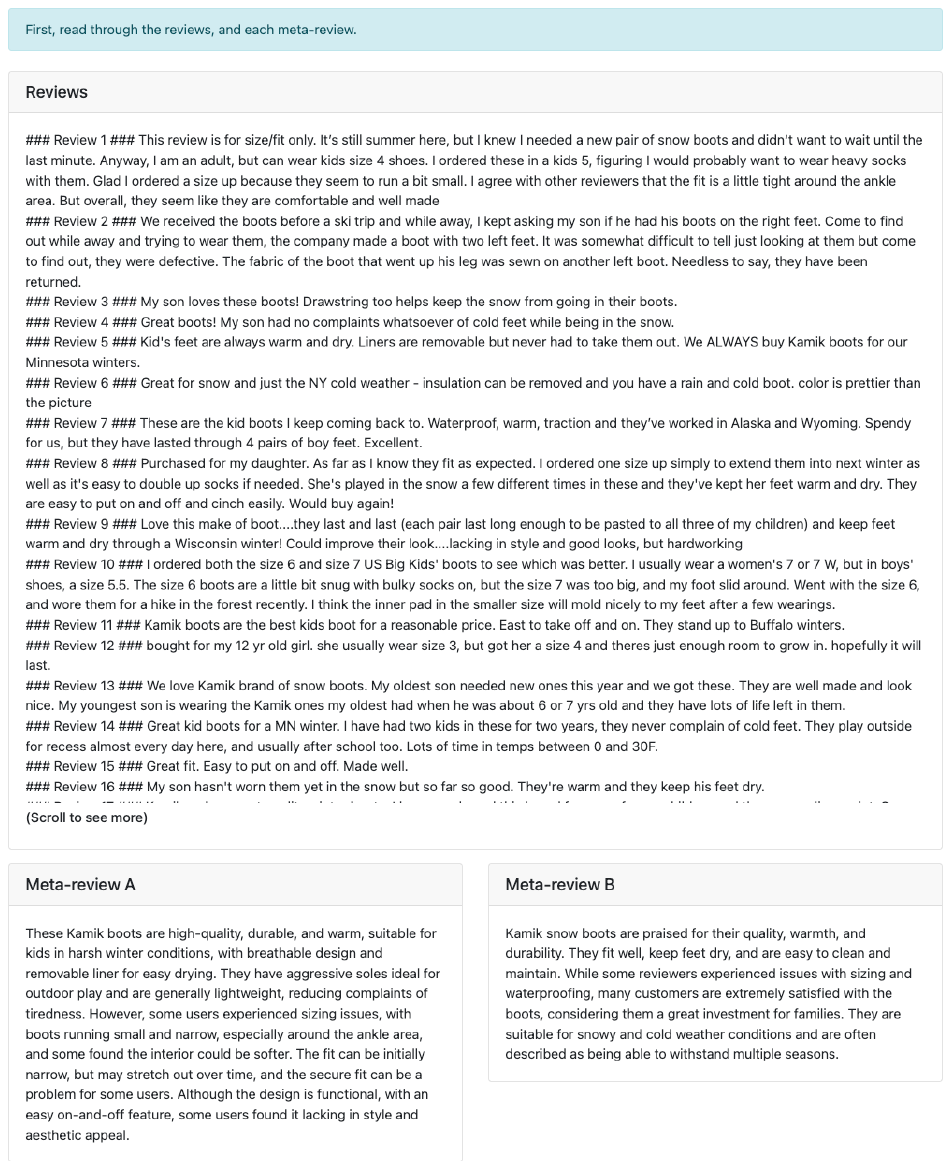}
\caption{Experimental instructions and interface for human evaluation
  study on sports shoes (part 2).}
\label{fig:meta_interface_shoes_p2}
\end{figure*}

\begin{figure*}[t]
\centering
\includegraphics[width=0.98\textwidth, trim=70 280 70 70, clip]{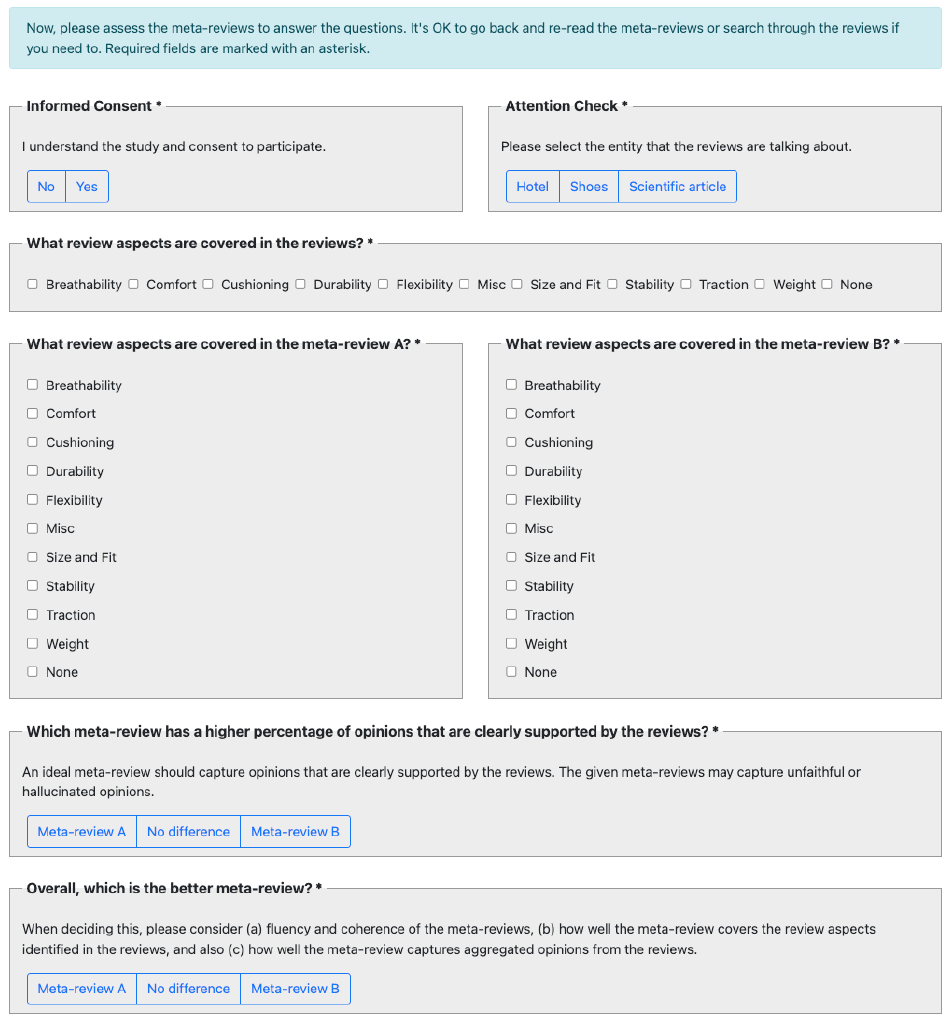}
\caption{Experimental instructions and interface for human evaluation
  study on sports shoes (part 3).}
\label{fig:meta_interface_shoes_p3}
\end{figure*}

We recruited 27 annotators from Prolific with L1 English from the US
or UK, with a minimum approval rate of 100\% in more than 100
studies. In addition to the attention check question for each annotation instance, we also included
quality control instances, asking participants to distinguish
human-written reference meta-reviews from random meta-reviews (taken from other entities). Each annotator worked on 20 annotation instances for the main study
and another 4 quality control instances.  Raters were asked five
questions about review aspects and opinion faithfulness. Our
annotation instructions and interface are shown
in~\figref{meta_interface_shoes_p1}, \figref{meta_interface_shoes_p2},
and \figref{meta_interface_shoes_p3}. After filtering out annotators
failing more than one quality control annotation pair, the annotators
have reasonable agreement and the average Krippendorff's $\alpha$
of~0.335.

\begin{figure*}[t]
\centering
\includegraphics[width=0.98\textwidth, trim=70 230 70 70, clip]{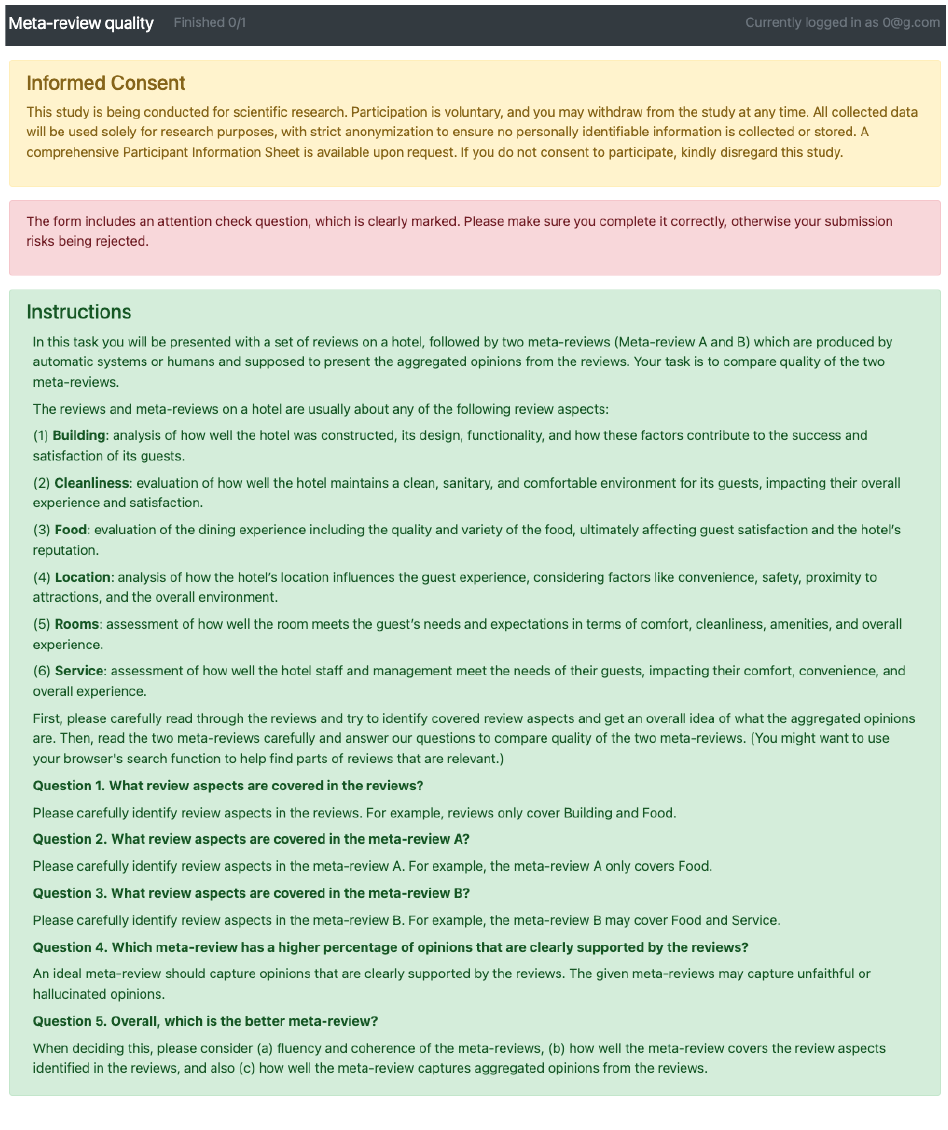}
\caption{Experimental instructions and interface for human evaluation
  study on hotels (part 1).}
\label{fig:meta_interface_hotel_p1}
\end{figure*}

\begin{figure*}[t]
\centering
\includegraphics[width=0.98\textwidth, trim=70 190 70 70, clip]{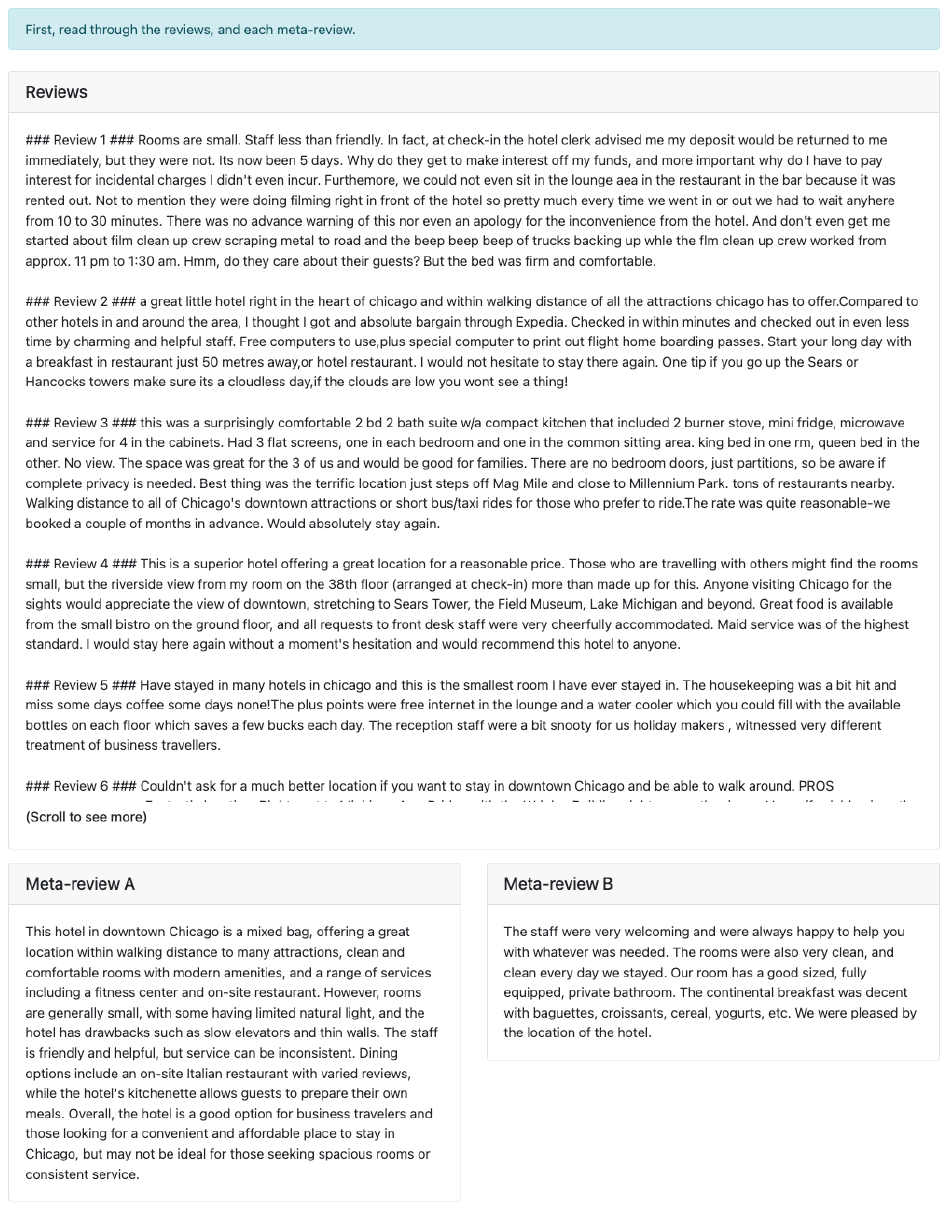}
\caption{Experimental instructions and interface for human evaluation
  study on hotels (part 2).}
\label{fig:meta_interface_hotel_p2}
\end{figure*}

\begin{figure*}[t]
\centering
\includegraphics[width=0.98\textwidth, trim=70 330 70 70, clip]{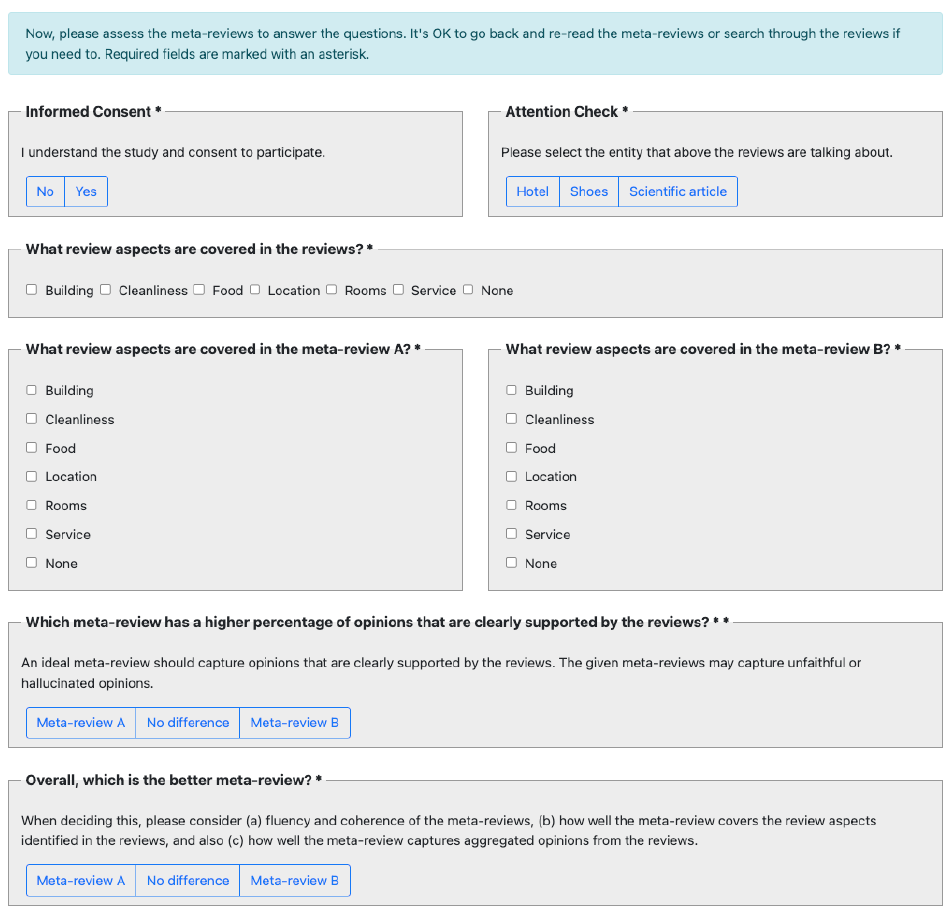}
\caption{Experimental instructions and interface for human evaluation
  study on hotels (part 3).}
\label{fig:meta_interface_hotel_p3}
\end{figure*}

We follow the same setting for the evaluation of meta-reviews for hotels.
There are also 540 annotations, and we obtain 27 annotators from
Prolific. The annotation instructions and experimental interface are
shown in~\figref{meta_interface_hotel_p1},
\figref{meta_interface_hotel_p2}, and
\figref{meta_interface_hotel_p3}. After filtering out annotators who
failed on more than one quality control instances, the average
Krippendorff's $\alpha$ is 0.622.

\begin{figure*}[ht!]
\centering
\includegraphics[width=0.98\textwidth, trim=70 240 70 70, clip]{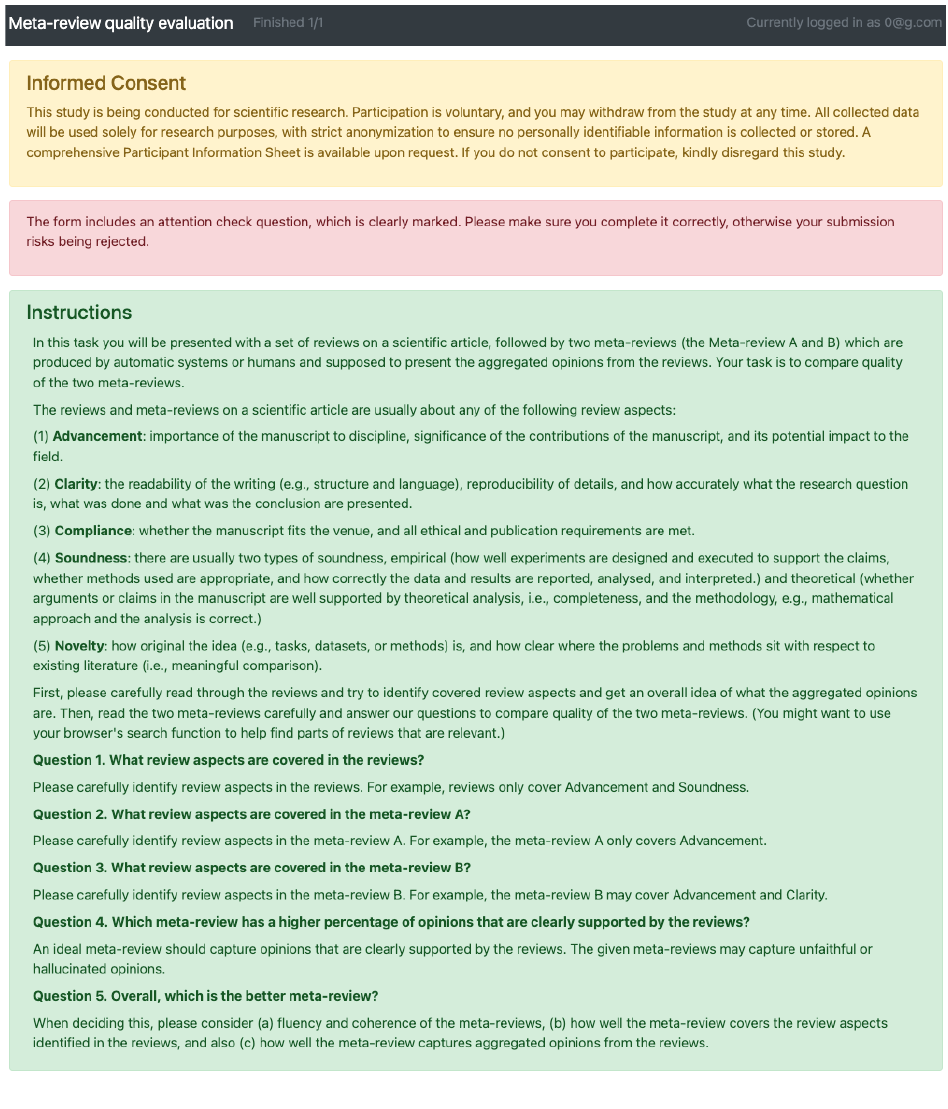}
\caption{Experimental instructions and interface for human evaluation
  study on article reviews (part 1).}
\label{fig:meta_interface_scientific_p1}
\end{figure*}

\begin{figure*}[t]
\centering
\includegraphics[width=0.98\textwidth, trim=70 180 70 70, clip]{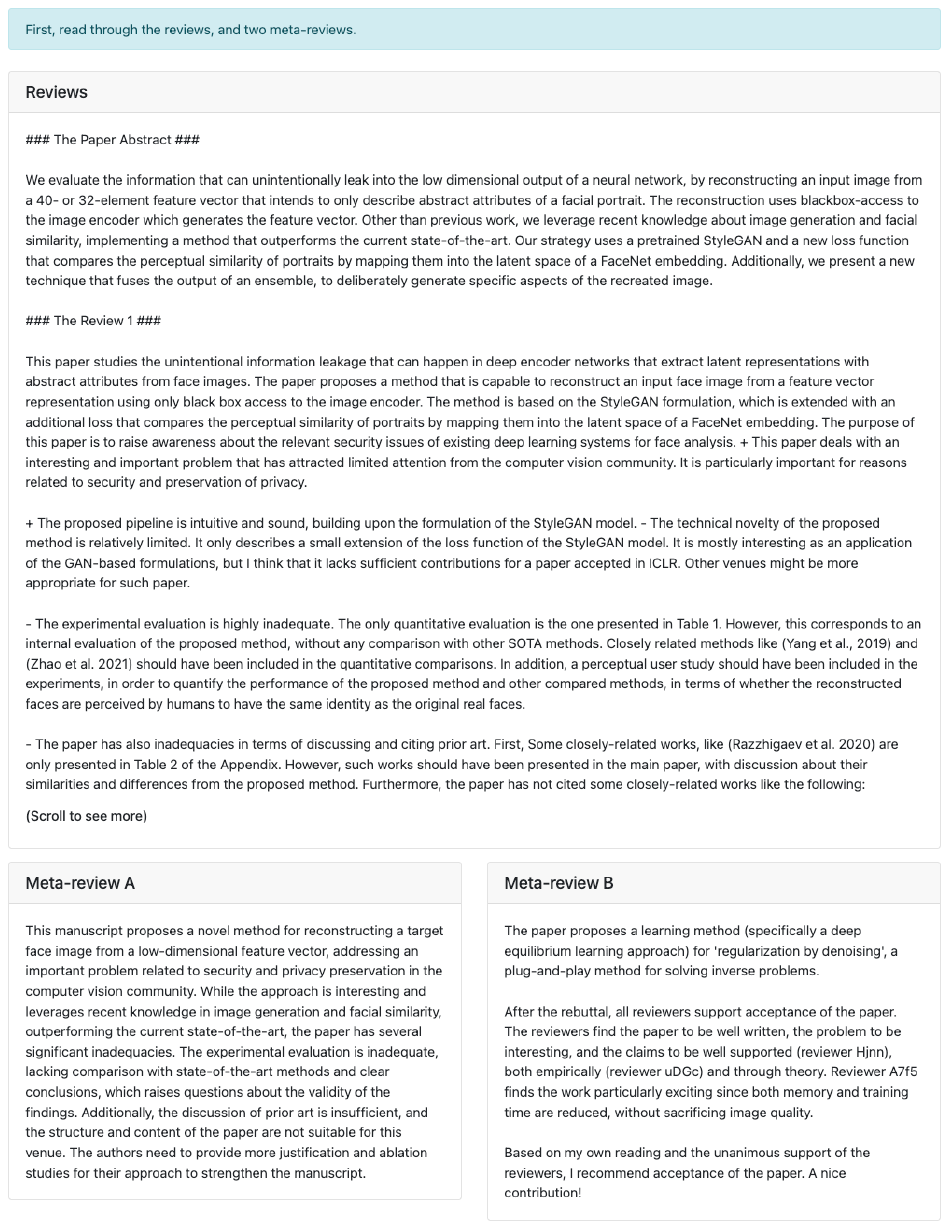}
\caption{Experimental instructions and interface for human evaluation
  study on article reviews (part 2).}
\label{fig:meta_interface_scientific_p2}
\end{figure*}

\begin{figure*}[t]
\centering
\includegraphics[width=0.98\textwidth, trim=70 350 70 70, clip]{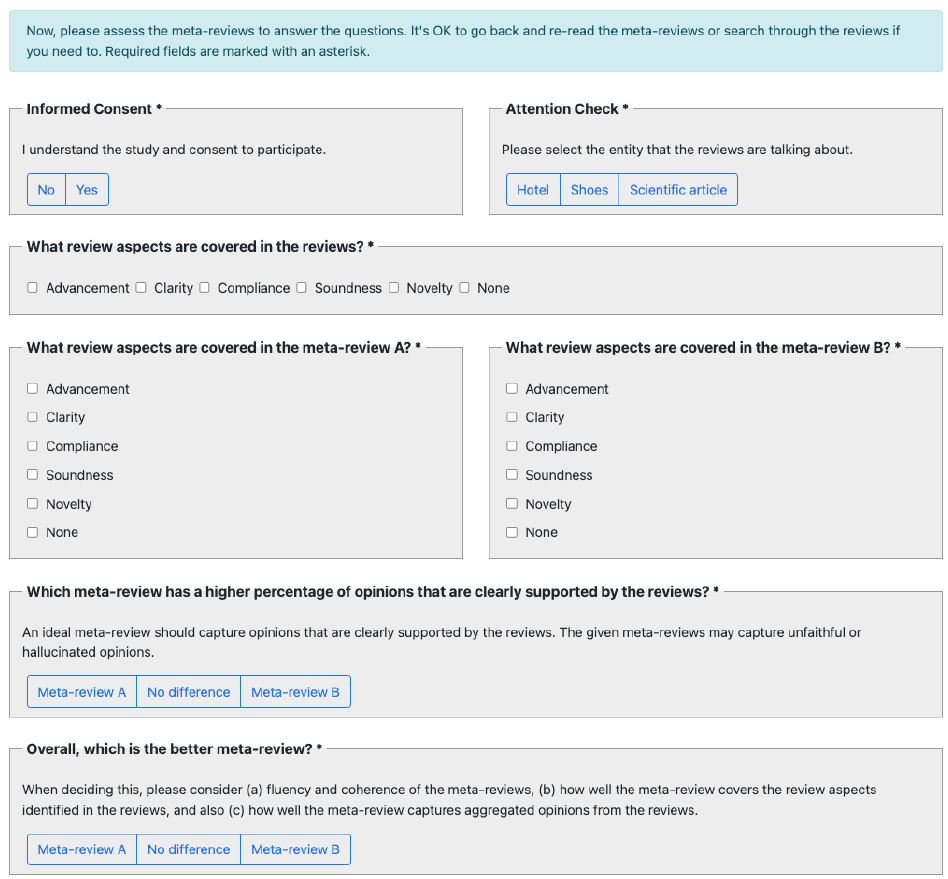}
\caption{Experimental instructions and interface for human evaluation
  study on article reviews (part 3).}
\label{fig:meta_interface_scientific_p3}
\end{figure*}

For scientific reviews of research articles, we randomly select ten
entities from the test data of PeerSum. There are also six pairs of
comparisons between our modular approach with Llama-3.1-70B as a
backbone and comparison baselines. As there are only about 15 reviews
on average, we show annotators all reviews. Therefore, there are 6
pairs of comparisons for each entity. Each pair gets annotated by
three different annotators and we have 180 annotations for the
dataset. We elicited 9 annotators from Prolific with required L1 English from
the US or UK, and a minimum approval rate of 100\% in more than 100
studies. We also required that they are pursuing a PhD in computer
science or engineering. In addition to the attention check question for each annotation instance, we
also included quality control instances, same as before. Therefore,
each annotator worked on 20 pairs of comparisons for the main study
and another 4 quality control instances. In each annotation,
participants are asked 5 questions about review aspects and opinion
faithfulness. The annotation instructions and interface are shown
in~\figref{meta_interface_scientific_p1},
\figref{meta_interface_scientific_p2}, and
\figref{meta_interface_scientific_p3}. After filtering out annotators
failing more than one quality control instances, the annotators, the
average Krippendorff's $\alpha$ is 0.463.

\section{Details of Human Evaluation on Usefulness of Intermediate Outputs}
\label{sec:details_human_intermediate}

\begin{figure*}[t]
\centering
\includegraphics[width=0.98\textwidth, trim=70 180 70 70, clip]{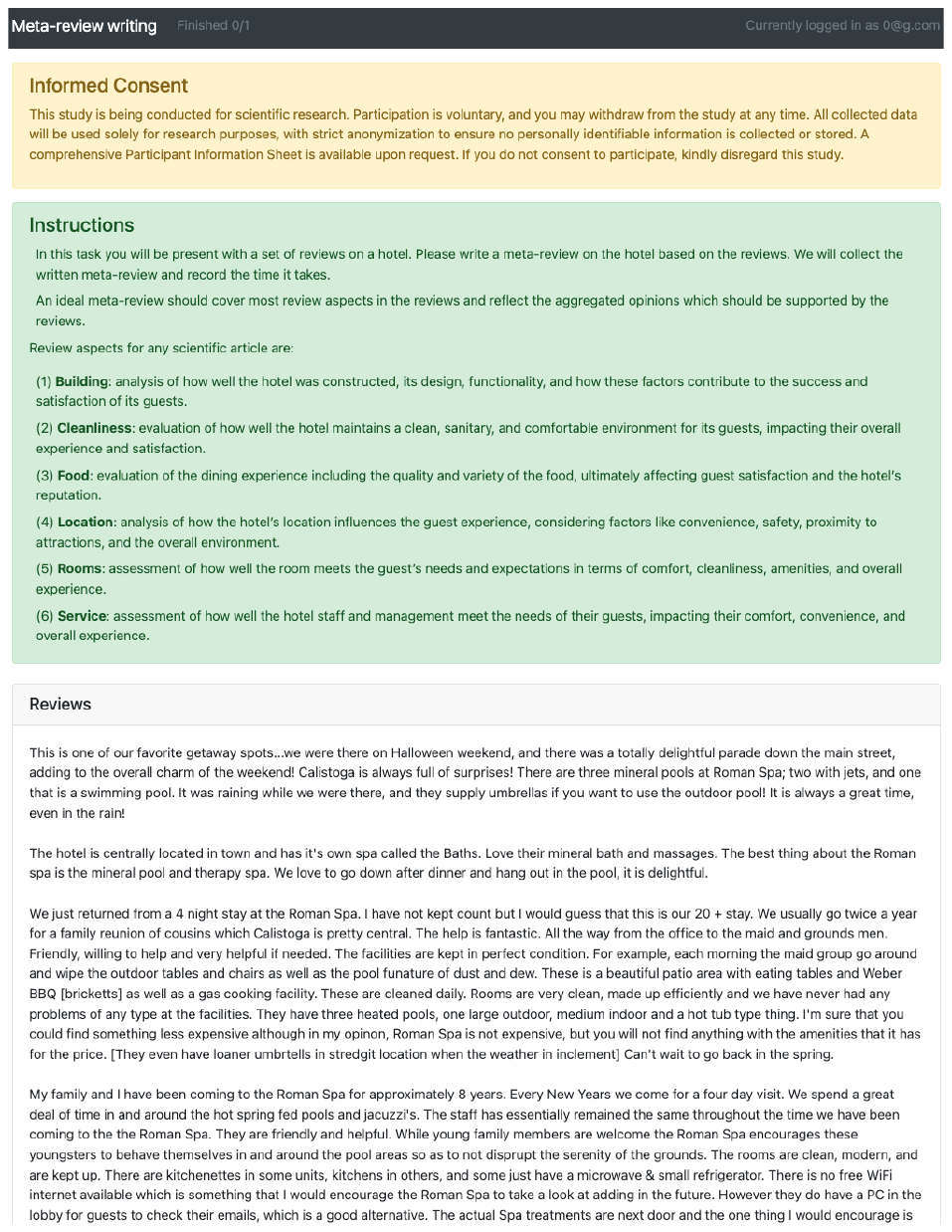}
\caption{Interface for  annotators to write meta-reviews based on
  different intermediate outputs (part 1).}
\label{fig:intermediate_interface_hotel_p1}
\end{figure*}

\begin{figure*}[t]
\centering
\includegraphics[width=0.98\textwidth, trim=70 180 70 70, clip]{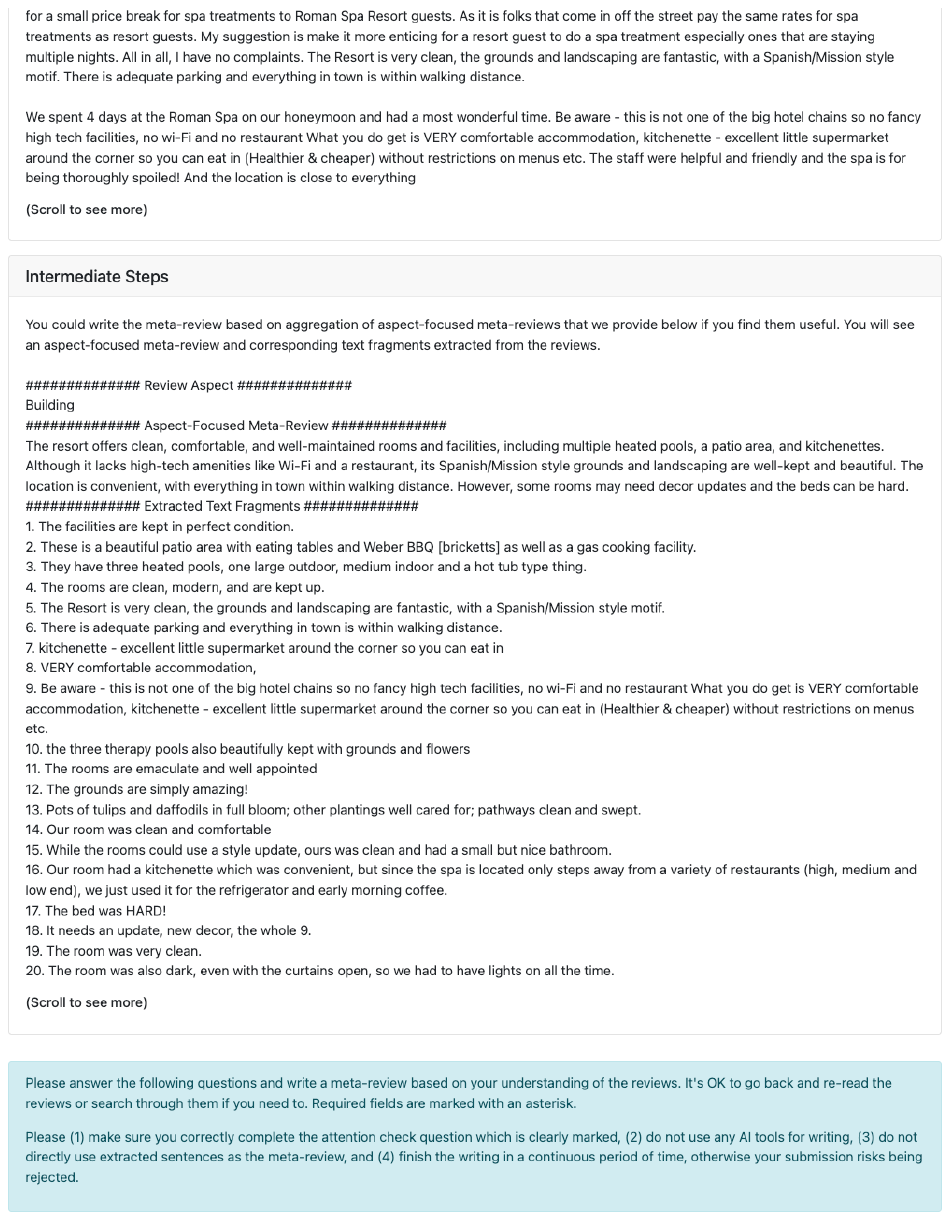}
\caption{Interface for annotators to write meta-reviews based on different intermediate outputs (part 2).}
\label{fig:intermediate_interface_hotel_p2}
\end{figure*}

\begin{figure*}[t]
\centering
\includegraphics[width=0.98\textwidth, trim=70 550 70 80, clip]{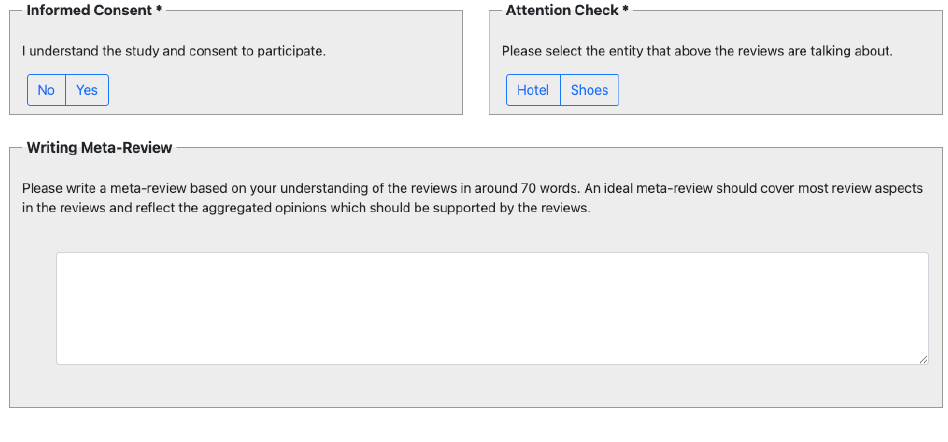}
\caption{Interface for  annotators to write meta-reviews based on different intermediate outputs (part 3).}
\label{fig:intermediate_interface_hotel_p3}
\end{figure*}

To record the time that humans spend to write meta-reviews with different reasoning steps, we conduct the experiments also with Prolific and present annotators interfaces with instructions in~\figref{intermediate_interface_hotel_p1}, \figref{intermediate_interface_hotel_p2} and \figref{intermediate_interface_hotel_p3}.We recruited five crowdworkers through
Prolific\footnote{www.prolific.com} with compensation above the UK
living wage at \pounds12 per working hour. These annotators are required to be experienced in L1 English from the US
or UK, with a minimum approval rate of 100\% in more than 100
studies. Annotators are required to focus on the annotation task and finish the writing task in a continuous period of time. The study is conducted on ten entities and there are three meta-reviews for each
(according to the three conditions described in~\secref{results_and_analysis}). To avoid memorization, each annotator must write a meta-review for each entity only once. We find that all our annotators passed our attention check question present in our instructions~\figref{intermediate_interface_hotel_p3}. We calculate the average time that the participants take for the ten instances in each condition from the five annotators.

To compare the quality of written meta-reviews in the three different conditions, we run another human evaluation in the same setting as the one to compare model-generated meta-reviews in \secref{results_and_analysis}. This was also based on pair-wise comparison and there were 30 pairs of comparison. We recruited three annotators and each pair of comparison was annotated for three times. The agreement among the three annotators is high (Krippendorff’s $\alpha$ is 0.542). 

\end{document}